\documentclass{article}

\usepackage{microtype}
\usepackage{graphicx}
\usepackage{subcaption}
\usepackage{booktabs} 
\usepackage{multirow}
\usepackage{xcolor}
\usepackage{colortbl}
\usepackage{makecell}
\usepackage{arydshln}
\usepackage{tcolorbox}
\tcbuselibrary{skins}
\usepackage{enumitem}

\usepackage{hyperref}

\usepackage[preprint]{icml2026}

\usepackage{amsmath}
\usepackage{amssymb}
\usepackage{mathtools}
\usepackage{amsthm}

\usepackage[capitalize,noabbrev]{cleveref}

\theoremstyle{plain}

\theoremstyle{definition}

\theoremstyle{remark}

\usepackage[textsize=tiny]{todonotes}

\icmltitlerunning{STAR : Bridging Statistical and Agentic Reasoning for Large Model Performance Prediction}

\begin{document}

\twocolumn[
  \icmltitle{STAR : Bridging Statistical and Agentic Reasoning for Large Model Performance Prediction}



  \icmlsetsymbol{equal}{*}
  \icmlsetsymbol{corr}{†}

\begin{icmlauthorlist}
    \icmlauthor{Xiaoxiao Wang}{equal,yyy,comp}
    \icmlauthor{Chunxiao Li}{equal,yyy,comp}
    \icmlauthor{Junying Wang}{yyy,comp}
    \icmlauthor{Yijin Guo}{sch,comp}
    \icmlauthor{Zijian Chen}{sch,comp}
    \icmlauthor{Chunyi Li}{sch,comp}
    \icmlauthor{Xiaohong Liu}{sch}
    \icmlauthor{Zicheng Zhang}{corr,sch,comp}
    \icmlauthor{Guangtao Zhai}{corr,sch,comp}
  \end{icmlauthorlist}

  \icmlaffiliation{yyy}{Fudan University}
  \icmlaffiliation{comp}{Shanghai Artificial Intelligence Laboratory}
  \icmlaffiliation{sch}{Shanghai Jiao Tong University}

  \icmlcorrespondingauthor{Zicheng Zhang, Guangtao Zhai}{wangxiaoxiao23@mails.ucas.ac.cn}

  \icmlkeywords{Machine Learning, ICML}

  \vskip 0.3in
]



\printAffiliationsAndNotice{}  


\begin{abstract}
As comprehensive large model evaluation becomes prohibitively expensive, predicting model performance from limited observations has become essential. However, existing statistical methods struggle with pattern shifts, data sparsity, and lack of explanation, while pure LLM methods remain unreliable. 
We propose \textbf{STAR}, a framework that bridges data-driven \textbf{ST}atistical expectations with knowledge-driven \textbf{A}gentic \textbf{R}easoning. 
STAR leverages specialized retrievers to gather external knowledge and embeds semantic features into Constrained Probabilistic Matrix Factorization (CPMF) to generate statistical expectations with uncertainty. 
A reasoning module guided by Expectation Violation Theory (EVT) then refines predictions through intra-family analysis, cross-model comparison, and credibility-aware aggregation, producing adjustments with traceable explanations.
Extensive experiments show that STAR consistently outperforms all baselines on both score-based and rank-based metrics, delivering a 14.46\% gain in total score over the strongest statistical method under extreme sparsity, with only 1--2 observed scores per test model.
\end{abstract}

\section{Introduction}
\begin{figure}
    \centering
    \includegraphics[width=\linewidth]{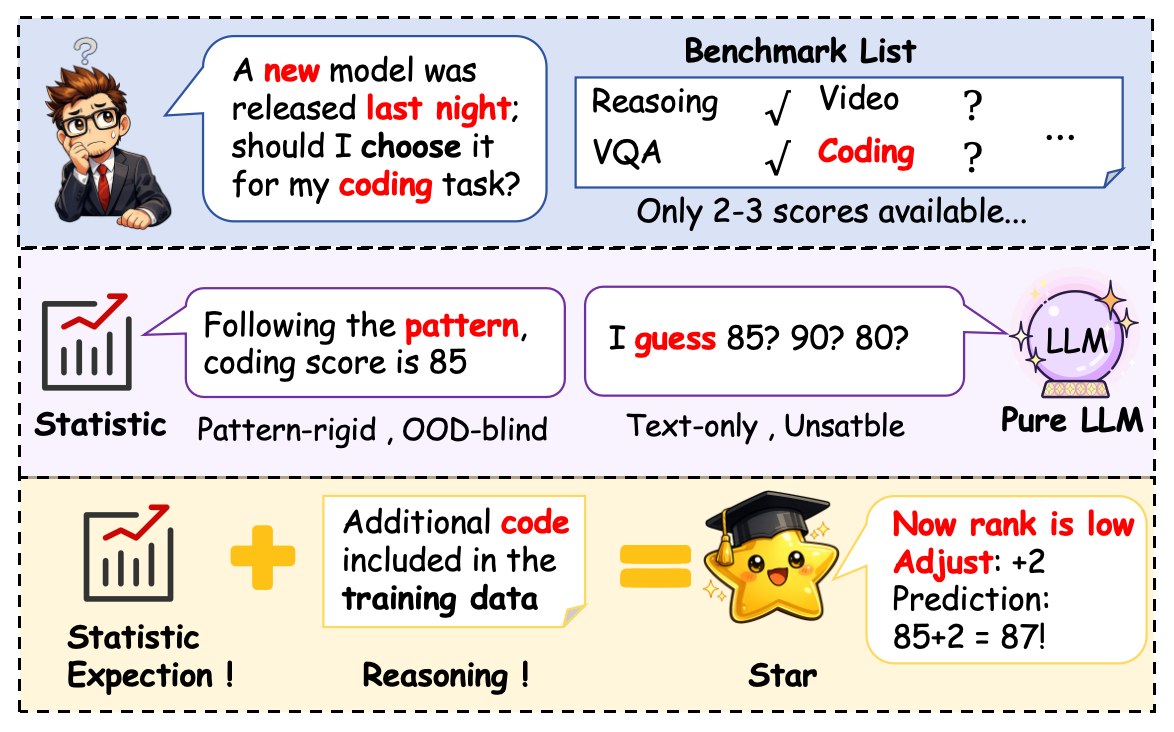}
    \caption{
    When predicting missing benchmark scores for new models, STAR combines statistical expectation with knowledge-driven reasoning to produce accurate and explainable predictions.}
    \label{fig:insight}
\end{figure}


The iteration of large models continues to accelerate, and evaluation benchmarks are also expanding rapidly \cite{AIBench,zhang2025large}.
Even evaluating a single LLM on HELM costs over 4,000 GPU hours or \$10,000 in API fees \citep{lee2023holistic}, and evaluation costs during model development can now exceed pre-training costs \citep{biderman2023pythia}. Critical decisions in AI deployment, including compute procurement\footnote{\href{https://ai.nowinstitute.org/publications/compute-and-ai}{AI Now Institute report}.}, regulatory compliance classification\footnote{\href{https://artificialintelligenceact.eu/implementation-timeline/}{EU AI Act implementation timeline}.}, and initial model selection\footnote{\href{https://coe.gsa.gov/coe/ai-guide-for-government}{U.S. government AI guide}.}, must be made before comprehensive evaluation is feasible. 
This evaluation bottleneck has motivated substantial research into efficient benchmarking methods, including subset selection approaches \cite{polo2024tinybenchmarks,perlitz2024efficient} that reduce per-benchmark cost but still require model access, and performance prediction techniques \cite{zhao2024can,zhang2024collaborative} that estimate performance across benchmarks from limited observations.

Previous performance prediction methods \cite{ruan2024observational} have demonstrated that large model performance can be characterized as a function of a low-dimensional capability space. The introduction of Scaling Laws \cite{kaplan2020scaling} has motivated approaches that employ Principal Component Analysis (PCA) \cite{ruan2024observational} or regression models \cite{polo2024sloth} to predict LLM performance based on factors such as parameter count, training data volume, and compute budget. Some works, on the other hand, formulate this as a matrix completion task, leveraging probabilistic matrix factorization (PMF) \cite{zhao2024can} or neural collaborative filtering (NCF) \cite{zhang2024collaborative} to model the performance function. However, compared to traditional statistical modeling scenarios, this problem poses three additional challenges: (1) \textbf{Pattern shifts}: The emergence of Mixture-of-Experts (MoE) architectures \cite{cai2025survey}, RLHF training paradigms \cite{ouyang2022training}, synthetic data \cite{li2023synthetic}, and so on continuously introduces 
out-of-distribution (OOD) patterns that invalidate historical correlations;
(2) \textbf{Extreme sparsity}: New models typically report scores on only a limited subset of benchmarks; and (3) \textbf{Explainability}: Practitioners require not merely a predicted score, but also evidence they can trust. Pure LLM methods \cite{park2025look} directly use LLMs to generate scores and explanations from model descriptions, but these methods suffer from severe hallucination issues.

To this end, we propose \textbf{STAR}, bridging data-driven \textbf{ST}atistical expectations and knowledge-driven \textbf{A}gentic \textbf{R}easoning in an explainable cognitive prediction framework. 
STAR maintains a memory module storing historical performance data alongside structured model and benchmark profiles. 
Specialized retrievers gather external knowledge from technical reports, model cards, benchmark specifications, and community feedback. 
Unlike traditional matrix completion methods that require manual feature engineering, STAR embeds retrieved semantic features into Constrained PMF (CPMF)~\cite{mnih2007probabilistic} to generate statistical expectations, while deriving prediction uncertainty via MCMC sampling~\cite{hoffman2014no}.

To refine statistical expectations, we design a reasoning module guided by Expectation Violation Theory (EVT)~\cite{burgoon1978communication, burgoon2015expectancy}. EVT describes a cognitive process in which humans form expectations based on past experience; when signals violate these expectations, they adjust their beliefs according to source credibility, emphasizing authoritative evidence while discounting unreliable signals.
We instantiate this principle by simulating how an expert AI practitioner gathers and evaluates evidence: (1) intra-family analysis: examining the model's positioning and architectural evolution within its family, (2) cross-model comparison: contrasting characteristics with rank-adjacent models, and (3) credibility-aware aggregation: synthesizing evidence by credibility to produce calibrated adjustments and traceable explanations. Our contributions include:
\begin{enumerate}
    \item We are the first to introduce EVT as a cognitive foundation for integrating statistical expectations with agentic reasoning in LLM performance prediction.
    \item We propose STAR, a retrieval-augmented CPMF framework with two-step credibility-aware reasoning, enabling explainable performance prediction.
    \item We conduct extensive experiments under extreme sparsity and pattern shifts, achieving a 14.46\% total score gain under 95\% masking and a 43.66 total improvement under benchmark-side shifts.
\end{enumerate}

\section{Related Work}
\subsection{Methods of Large Model Benchmark Prediction}
\begin{table}[th]
\setlength{\tabcolsep}{2pt}
\centering
\small
\begin{tabular}{lcccc}
\toprule
\textbf{Method}  & \textbf{Statistics} & \textbf{Reasoning} & \textbf{Retrieval}\\
\midrule
PCA \cite{ruan2024observational} & \textcolor{green!70!black}{$\checkmark$} & \textcolor{red}{$\times$} & \textcolor{red}{$\times$}\\
Regression \cite{polo2024sloth} & \textcolor{green!70!black}{$\checkmark$} & \textcolor{red}{$\times$} & \textcolor{red}{$\times$}\\
PMF~\cite{zhao2024can} & \textcolor{green!70!black}{$\checkmark$} & \textcolor{red}{$\times$} & \textcolor{red}{$\times$}\\
NCF~\cite{zhang2024collaborative} & \textcolor{green!70!black}{$\checkmark$} & \textcolor{red}{$\times$} & \textcolor{red}{$\times$}\\
Pure LLM~\cite{park2025look} & \textcolor{red}{$\times$} & \textcolor{green!70!black}{$\checkmark$} & \textcolor{red}{$\times$}\\
\midrule
\textbf{ours}& \textcolor{green!70!black}{$\checkmark$} & \textcolor{green!70!black}{$\checkmark$} & \textcolor{green!70!black}{$\checkmark$}\\
\bottomrule
\end{tabular}
\caption{Comparison of performance prediction approaches for large models.}
\label{tab:method_comparison}
\end{table}

The redundancy of benchmarks~\cite{zhang2025redundancy,wen2025improve} and rapid release of new models make performance prediction increasingly valuable for reducing evaluation costs. Scaling-law approaches such as PCA-based capability extraction~\cite{ruan2024observational} and latent skill modeling with family-specific efficiency~\cite{polo2024sloth} provide low-dimensional estimates of model performance. Another line of work frames prediction as a recommender system problem, applying PMF~\cite{zhao2024can} or NCF~\cite{zhang2024collaborative} to learn low-rank performance representations.
While these statistical methods reveal inherent predictability, they are less robust under pattern shifts and limited by the lack of external knowledge for novel models. Alternatively, LLM prompting with model descriptions~\cite{park2025look} ignores statistical signals and offers no uncertainty control. Our method bridges both paradigms by combining statistical expectations with LLM-based semantic reasoning, achieving robustness to novel models while providing traceable explanations. Table~\ref{tab:method_comparison} summarizes the key differences.

\begin{figure*}[t]
    \centering
    \includegraphics[width=\linewidth]{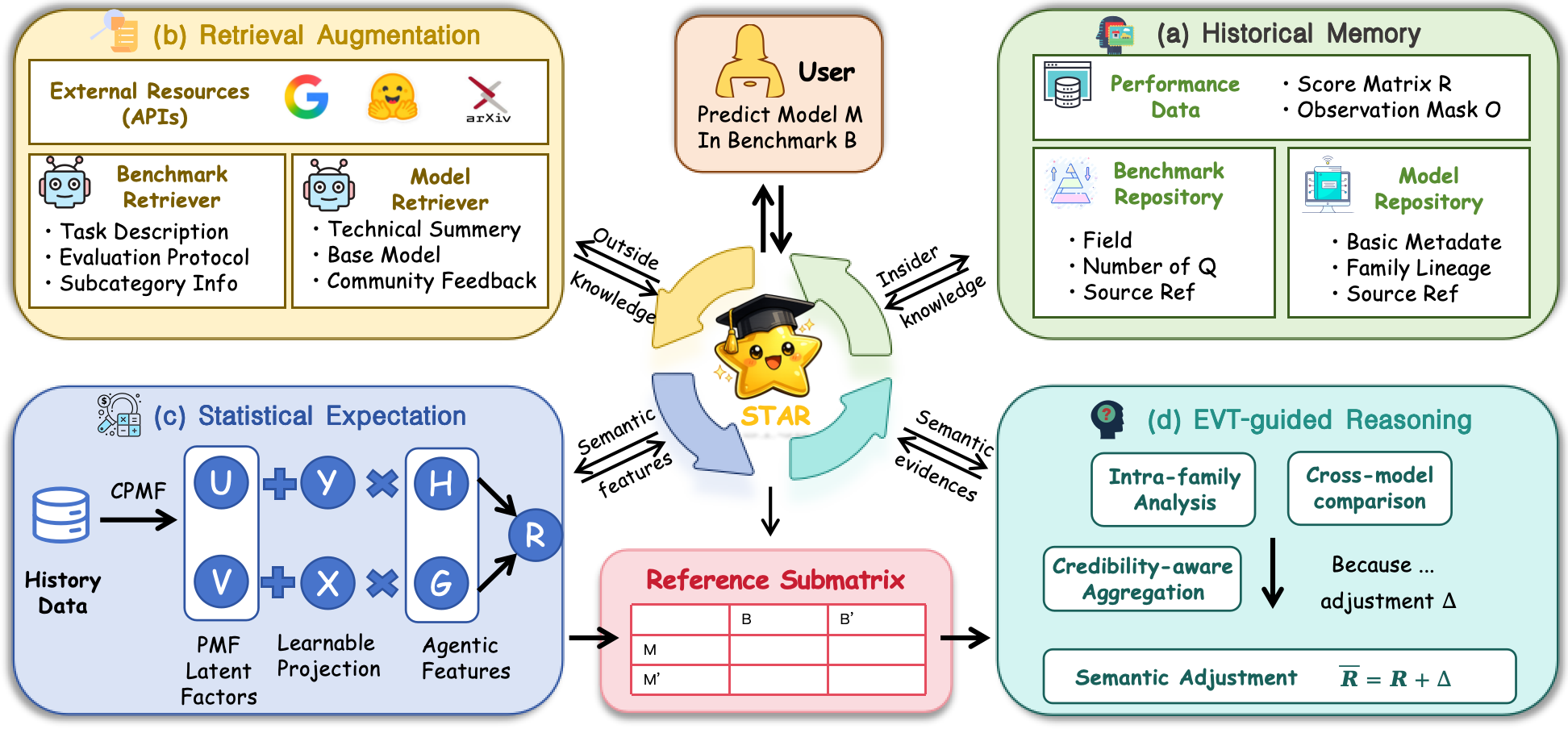}
    \caption{\textbf{Overview of the STAR framework.} 
 Given a user query to predict model $M$'s performance on benchmark $B$, STAR integrates four modules: (a) \textbf{Historical Memory} maintains the observed score matrix and structured profiles for models and benchmarks; (b) \textbf{Retrieval Augmentation} collects external knowledge from google, huggingface and arxiv; (c) \textbf{Statistical Expectation} combines PMF latent factors with retrieval semantic features to generate initial expectations with uncertainty estimates; (d) \textbf{EVT-guided Reasoning} refines predictions through two-step analysis and credibility-aware aggregation, producing adjustments $\Delta$ with explanations. 
}
    \label{fig:framework}
\end{figure*}

\subsection{LLM-based Agents for Prediction}

LLM-based agents have been applied as versatile prediction systems across diverse domains. In finance, FINCON~\cite{yu2024fincon} employs manager-analyst hierarchies for risk-aware trading, while FinMem~\cite{yu2025finmem} implements layered memory mimicking human cognition. For time series, Time-LLM~\cite{jin2023time} adapts frozen LLM backbones via textual reprogramming. In healthcare, AgentMD~\cite{jin2025agentmd} and MDAgents~\cite{kim2024mdagents} coordinate clinical tools through adaptive collaboration, while scientific agents such as ChemCrow~\cite{m2024augmenting} integrate expert tools with ReAct-style reasoning. Collectively, these works highlight recurring patterns including memory augmentation, tool orchestration, and multi-agent collaboration.
However, most designs remain engineering-driven, lacking principled guidance on how to integrate statistical signals with external knowledge. STAR addresses this gap by grounding evidence integration in Expectation Violation Theory (EVT).


\section{Method}

\subsection{Problem Formulation}

Let $\mathbf{R} \in \mathbb{R}^{M \times N}$ be the performance matrix of $M$ models on $N$ benchmarks, with $R_{mn}$ denoting the score of model $m$ on benchmark $n$. Only a sparse subset is observed in practice, indicated by $\mathbf{O} \in \{0,1\}^{M \times N}$ where $O_{mn}=1$ if $R_{mn}$ is known. Let $\mathcal{I}_m$ and $\mathcal{I}_n$ denote auxiliary knowledge for model $m$ and benchmark $n$, respectively, such as reports or task descriptions. Given $\mathbf{R}^{\mathrm{obs}}=\mathbf{O}\odot\mathbf{R}$ and knowledge $\{\mathcal{I}_m\}_{m=1}^M$, $\{\mathcal{I}_n\}_{n=1}^N$, our goals are to predict missing scores $\hat{R}_{mn}$ for $O_{mn}=0$, and generate explanations $\mathcal{E}_{mn}$ that justify each prediction with traceable evidence.

\subsection{Historical Memory}
The historical memory module serves as the foundational knowledge base of STAR, maintaining both quantitative performance records and structured semantic profiles. As illustrated in Figure~\ref{fig:framework} (a), this module comprises three interconnected components: (1) Performance Data Storage, which maintains the observed score matrix $\mathbf{R}^{\mathrm{obs}} = \mathbf{O} \odot \mathbf{R}$; (2) Model Repository, which stores structured profiles $\mathcal{P}_m$ for each model including basic metadata, family lineage, and source references; and (3) Benchmark Repository, which maintains profiles $\mathcal{Q}_n$ containing field, number of questions and source references. We denote the complete memory state as $\mathcal{M} = \{\mathbf{R}^{\mathrm{obs}}, \mathbf{O}, \{\mathcal{P}_m\}_{m=1}^M, \{\mathcal{Q}_n\}_{n=1}^N\}$, supporting both retrieval and reasoning modules.
\subsection{Retrieval Augmentation}
\label{sec:retrieval}
While the historical memory module provides structured profiles, it may lack external knowledge for accurate prediction. To address this, we design two specialized retrievers that collect information from search engines, arXiv, and HuggingFace (Figure~\ref{fig:framework}(b)).

The benchmark retriever gathers task-specific information for benchmark $n$, including task descriptions, evaluation protocols, and subcategory description. The model retriever collects evidence for model $m$ following a structured schema: (1) technical summary, covering model introduction, key enhancements over predecessors, and architectural updates; (2) base model analysis of the underlying language model or vision-language model that the target model builds upon; and (3) community feedback, capturing empirical evaluations, practitioner reviews, and technical reports when official documentation is limited. To prevent information leakage, we constrain retrieval to a short window (within 8 days) after the model's release date, which sufficient to capture official documentation while excluding subsequent evaluation reports, and explicitly filter out any benchmark performance scores from the collected evidence.
These retrieved texts serve dual purposes: they are encoded into semantic feature vectors $\mathbf{h}_n, \mathbf{g}_m \in \mathbb{R}^{d}$ via a text encoder for the statistical expectation module (\S\ref{sec:statistical}), and preserved as raw textual evidence $\mathcal{T}_n, \mathcal{T}_m$ to support interpretable reasoning in the EVT-guided reasoning module (\S\ref{sec:evt}).

\subsection{Statistical Expectation}
\label{sec:statistical}

\textbf{Probabilistic Matrix Factorization (PMF).}
We build upon PMF~\cite{mnih2007probabilistic} to model the performance matrix. PMF decomposes $\mathbf{R}$ into two low-rank matrices $\mathbf{U} \in \mathbb{R}^{M \times D}$ and $\mathbf{V} \in \mathbb{R}^{N \times D}$, where $D$ is the latent dimension. We place Gaussian priors on the latent vectors:
\begin{equation}
\begin{split}
    p(\mathbf{U}) &= \prod_{m=1}^{M} \mathcal{N}(\mathbf{U}_m \mid \mathbf{0}, \sigma_U^2 \mathbf{I}), \\
    p(\mathbf{V}) &= \prod_{n=1}^{N} \mathcal{N}(\mathbf{V}_n \mid \mathbf{0}, \sigma_V^2 \mathbf{I}),
\end{split}
\end{equation}
and model the observed scores with Gaussian likelihood:
\begin{equation}
    p(\mathbf{R} \mid \mathbf{U}, \mathbf{V}, \sigma^2) = \prod_{(m,n): O_{mn}=1} \mathcal{N}(R_{mn} \mid \mathbf{U}_m^\top \mathbf{V}_n, \sigma^2).
\end{equation}

\textbf{Semantic-Enhanced Constrained PMF.}
Constrained PMF (CPMF)~\cite{mnih2007probabilistic} augments latent factors with auxiliary features to address cold-start scenarios. \citet{zhao2024can} adapt this idea by incorporating manually designed discrete features such as one-hot encodings of vision encoder types and model families. However, such representations require domain expertise for each new attribute and fail to capture nuanced semantic information \cite{bengio2013representation,mikolov2013distributed}. We instead leverage the dense semantic features $\mathbf{g}_m$ and $\mathbf{h}_n$ from our retrieval module (\S\ref{sec:retrieval}). To integrate these features into the latent space, we introduce learnable projection matrices $\mathbf{X} \in \mathbb{R}^{d \times D}$ and $\mathbf{Y} \in \mathbb{R}^{d \times D}$ with Gaussian priors:
\begin{equation}
    \mathbf{U}'_m = \mathbf{U}_m + \mathbf{g}_m \mathbf{X}, \quad \mathbf{V}'_n = \mathbf{V}_n + \mathbf{h}_n \mathbf{Y}.
\end{equation}

The statistical expectation is computed as $\hat{R}_{mn} = {\mathbf{U}'_m}^\top \mathbf{V}'_n$.
This formulation offers two key advantages over discrete handcrafted features. 
First, semantic embeddings inject rich knowledge from retrieved technical documentation, enabling CPMF to model fine-grained differences beyond coarse categorical attributes. 
Second, because these features are automatically derived from open-world text, the approach generalizes naturally to newly released models and benchmarks without manual feature construction.

\textbf{Uncertainty Estimation.}
We apply Markov Chain Monte Carlo (MCMC) sampling using the No-U-Turn Sampler (NUTS)~\cite{hoffman2014no} to obtain posterior distributions over latent variables. Given $T$ posterior samples $\{(\mathbf{U}^{(t)}, \mathbf{V}^{(t)}, \mathbf{X}^{(t)}, \mathbf{Y}^{(t)})\}_{t=1}^{T}$, we compute the mean prediction and uncertainty:
\begin{equation}
\begin{split}
    \hat{R}_{mn} &= \frac{1}{T} \sum_{t=1}^{T} {\mathbf{U}'^{(t)}_m}^\top \mathbf{V}'^{(t)}_n, \\
    \sigma_{mn} &= \sqrt{\mathrm{Var}\left( \{ {\mathbf{U}'^{(t)}_m}^\top \mathbf{V}'^{(t)}_n \}_{t=1}^{T} \right)}.
\end{split}
\end{equation}
The uncertainty $\sigma_{mn}$ quantifies the model's confidence in its statistical prediction. This uncertainty estimate informs the subsequent EVT-guided reasoning module (\S\ref{sec:evt}): when $\sigma_{mn}$ is high, the statistical expectation alone may be unreliable, signaling that the model should rely more heavily on semantic evidence to calibrate its prediction.

\begin{table*}[htbp]
\centering
\small
\setlength{\tabcolsep}{4.5pt}
\begin{tabular}{lcccccccc}
\toprule
 \multirow{2}{*}{Method} & \multicolumn{3}{c}{\textbf{Score-based Metrics$\downarrow$}} & \multicolumn{4}{c}{\textbf{Rank-based Metrics$\uparrow$}} & \multicolumn{1}{c}{\textbf{Total$\uparrow$}} \\
\cmidrule(lr){2-4} \cmidrule(lr){5-8} \cmidrule(lr){9-9}
& RMSE & MAE & Score Avg & SRCC(\%) & KRCC(\%) & MAE@3(\%) & Rank Avg(\%) & Score \\
\midrule
\rowcolor{gray!15}\multicolumn{9}{l}{\textit{Statistical Baselines}} \\
 
Global Mean & 16.19 & 13.01 & 14.60 & 85.61 & 77.02 & 10.75 & 57.79 & 43.19 \\
Mean Of Means & 12.93 & 9.74 & 11.33 & 92.28 & 83.76 & 33.74 & 69.93 & 58.60 \\
\midrule
\rowcolor{gray!15}\multicolumn{9}{l}{\textit{Statistic-based Methods}} \\
PCA~\cite{ruan2024observational} & 15.33 & 11.82 & 13.58 & 86.86 & 78.04 & 14.99 & 59.96 & 46.38 {\scriptsize $\pm$ 0.85} \\
Regression~\cite{polo2024sloth} & 13.01 & 8.98 & 11.00 & 91.70 & 83.58 & \underline{38.04} & 71.11 & 60.11 {\scriptsize $\pm$ 0.72} \\
NCF~\cite{zhang2024collaborative} & 13.13 & 8.95 & 11.04 & 90.44 & 81.80 & 22.45 & 64.90 & 53.86 {\scriptsize $\pm$ 0.88} \\
PMF~\cite{zhao2024can} & 12.82 & 8.18 & 10.50 & 92.46 & 84.42 & 30.48 & 69.12 & 58.62 {\scriptsize $\pm$ 0.68} \\
\midrule
\rowcolor{gray!15}\multicolumn{9}{l}{\textit{LLM-based Methods}} \\
Pure LLM~\cite{park2025look} & 23.33 & 16.55 & 19.94 & 70.60 & 65.30 & 24.10 & 53.33 & 33.39 {\scriptsize $\pm$ 1.25} \\
LLM-RAG & 14.64 & 9.25 & 11.94 & 76.65 & 61.06 & 27.35 & 55.02 & 43.08 {\scriptsize $\pm$ 1.12} \\
\midrule
Semantic-Augmented CPMF & \underline{9.95} & \underline{6.47} & \underline{8.21} & \underline{95.27} & \underline{87.56} & 33.94 & \underline{72.26} & \underline{64.05} {\scriptsize $\pm$ 0.72} \\
STAR & \textbf{8.75} & \textbf{5.69} & \textbf{7.22} & \textbf{96.10} & \textbf{88.64} & \textbf{38.23} & \textbf{74.32} & \textbf{67.10} {\scriptsize $\pm$ 0.82} \\
\bottomrule
\end{tabular}
\caption{Performance comparison under 95\% masking. Results are reported as mean {\scriptsize $\pm$ std} over 5 runs. Lower is better for score-based metrics; higher is better for rank-based metrics and total score. \textbf{Bold} indicates best results; \underline{underline} indicates second best.}
\label{tab:model_comparison_95}
\end{table*}

\subsection{EVT-guided Reasoning}
\label{sec:evt}

Statistical expectations from CPMF may be unreliable under limited observations or pattern shifts. Inspired by Expectation Violation Theory (EVT)~\cite{burgoon1978communication,afifi2000impact,burgoon2015expectancy}: humans adjust expectations based on evidence credibility. We design a reasoning module that refines predictions through semantic analysis. As shown in Figure~\ref{fig:framework}(d), given $\hat{R}_{mn}$, $\sigma_{mn}$, and retrieved evidence $\mathcal{T}_m, \mathcal{T}_n$, the module performs two-step analysis followed by credibility-aware aggregation.

\textbf{Step 1: Intra-family Analysis.}
The first step examines the target model's positioning within its model family $\mathcal{F}_m$ (e.g., LLaVA series, InternVL series). The reasoning agent takes the statistical expectation, retrieved evidence, and family members' observed scores as input:
\begin{equation}
    \mathcal{A}_1 = f_{\text{LLM}}\left( \hat{R}_{mn}, \mathcal{T}_m, \{R_{m'n}\}_{m' \in \mathcal{F}_m} \right)
\end{equation}
This analysis identifies reference scores, examines architectural evolution, and assesses whether specific optimizations may affect performance on benchmark $n$.

\textbf{Step 2: Cross-model Comparison.}
The second step compares the target model against capability-similar models $\mathcal{C}_m$, identified by performance similarity on shared benchmarks. The reasoning agent analyzes:
\begin{equation}
    \mathcal{A}_2 = f_{\text{LLM}}\left( \hat{R}_{mn}, \{R_{m'n}\}_{m' \in \mathcal{C}_m} \right)
\end{equation}
This comparison reveals whether the statistical expectation is consistent with similar models' actual performance, helping detect potential over- or under-estimation.

\textbf{Credibility-aware Aggregation.}
The reasoning agent synthesizes findings from both analysis steps, producing an adjustment $\Delta_{mn}$, confidence score $c_{mn} \in [0, 1]$, and natural language explanation $\mathcal{E}_{mn}$. Following EVT principles, the agent assigns higher confidence when: (1) multiple sources converge on consistent conclusions, (2) direct family references are available, or (3) statistical uncertainty $\sigma_{mn}$ is high, indicating room for semantic-based correction. The final prediction combines the statistical expectation with the credibility-weighted adjustment:
\begin{equation}
    \tilde{R}_{mn} = \hat{R}_{mn} + c_{mn} \cdot \Delta_{mn}
\end{equation}
This formulation embodies EVT: authoritative evidence triggers larger corrections, while weak or conflicting evidence results in conservative adjustments.

\section{Experiment}
\begin{figure*}[h]
    \centering
    \includegraphics[width=\linewidth]{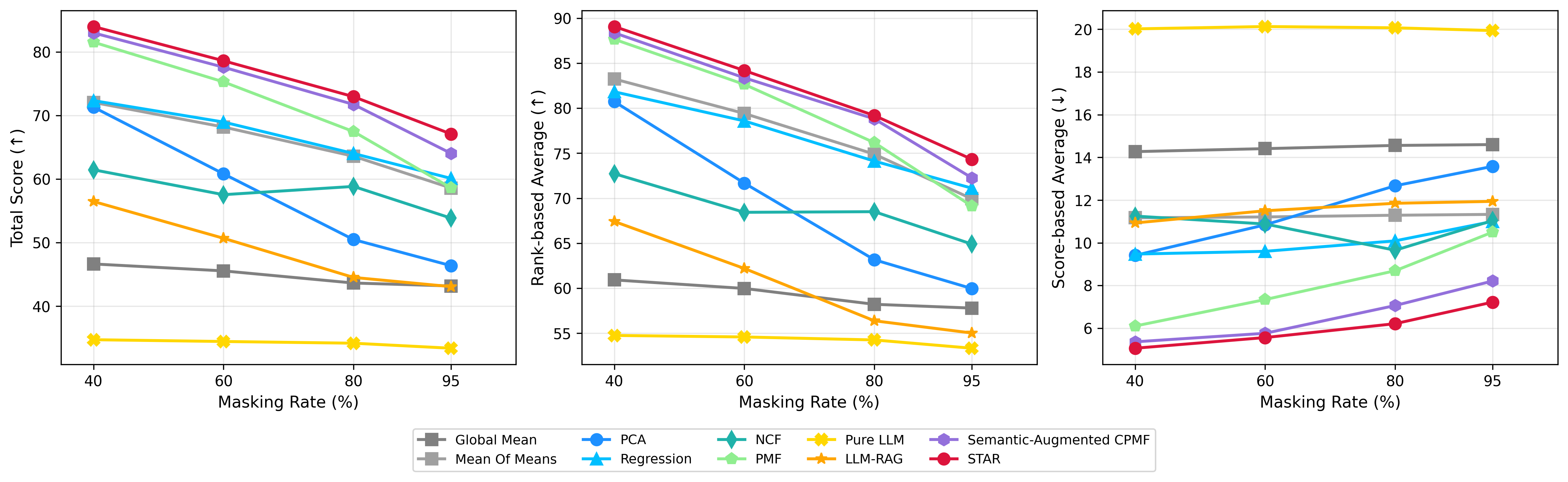}
    \caption{Performance comparison across masking rates from 40\% to 95\%. Left: Total Score (Rank Avg $-$ Score Avg, higher is better). Middle: Rank-based Average (higher is better). Right: Score-based Average (lower is better). STAR consistently achieves the best performance across all metrics, with its advantage over statistical baselines widening as sparsity increases. Details in Appendix \ref{app:sparsity_result}}
    \label{fig:masking rate}
\end{figure*}
\subsection{Experimental Setup}

\textbf{Dataset.} We construct our evaluation benchmark using data from OpenCompass\footnote{\href{http://opencompass.openxlab.space/assets/OpenVLM.json}{OpenCompass}.}, a comprehensive platform for evaluating LLMs and VLMs. The performance matrix comprises 285 models across 28 benchmarks, with 6,032 valid entries spanning visual reasoning, OCR, chart understanding, and multimodal knowledge. See Appendix \ref{app:benchmark_description} for details.

\textbf{Evaluation Metrics.} We evaluate using eight metrics organized into two categories. Score-based metrics include MAE and RMSE, which measure prediction accuracy on scores normalized to the 0--100 scale, along with their average termed Score Avg. Rank-based metrics include SRCC, KRCC, and MAE@3, which assess ranking consistency and are reported as percentages, along with their average termed Rank Avg. We introduce Total Score, computed as Rank Avg $-$ Score Avg, as a unified indicator that balances both prediction accuracy and ranking quality. Detailed definitions are provided in Appendix~\ref{sec:metrics}.

\textbf{Baselines.} We compare STAR against two categories of methods. Statistical-based methods: Global Mean, which predicts the global average score across all observed entries; Mean of Means, which averages row means (model-wise) and column means (benchmark-wise); PCA~\cite{ruan2024observational}; Regression~\cite{polo2024sloth}; NCF~\cite{zhang2024collaborative}; and PMF~\cite{zhao2024can}. LLM-based methods include: Pure LLM~\cite{park2025look}, which generates predictions based solely on structured benchmark descriptions; and LLM-RAG, which combines retrieval with LLM reasoning but without CPMF.

\textbf{Implementation Details.} 
We adopt GPT-5.1 as the backbone LLM for retrieval and reasoning modules, with web search disabled and a knowledge cutoff of September 30, 2024. To prevent information leakage, we implement filtering mechanisms that block any benchmark score-related content from being included in the retrieved evidence. For semantic feature encoding, we employ BGE-M3~\cite{xiao2024c} to obtain $d=1024$ dimensional representations for both models and benchmarks. In the CPMF module, we set the latent dimension to $D=10$ and perform posterior inference via MCMC sampling with $T=100$ samples following a burn-in period of 100 iterations.

\subsection{Estimating Unknown Performance}

\subsubsection{Prediction for Sparsity Performance}
\begin{figure*}[h]
    \centering
    \includegraphics[width=0.9\linewidth]{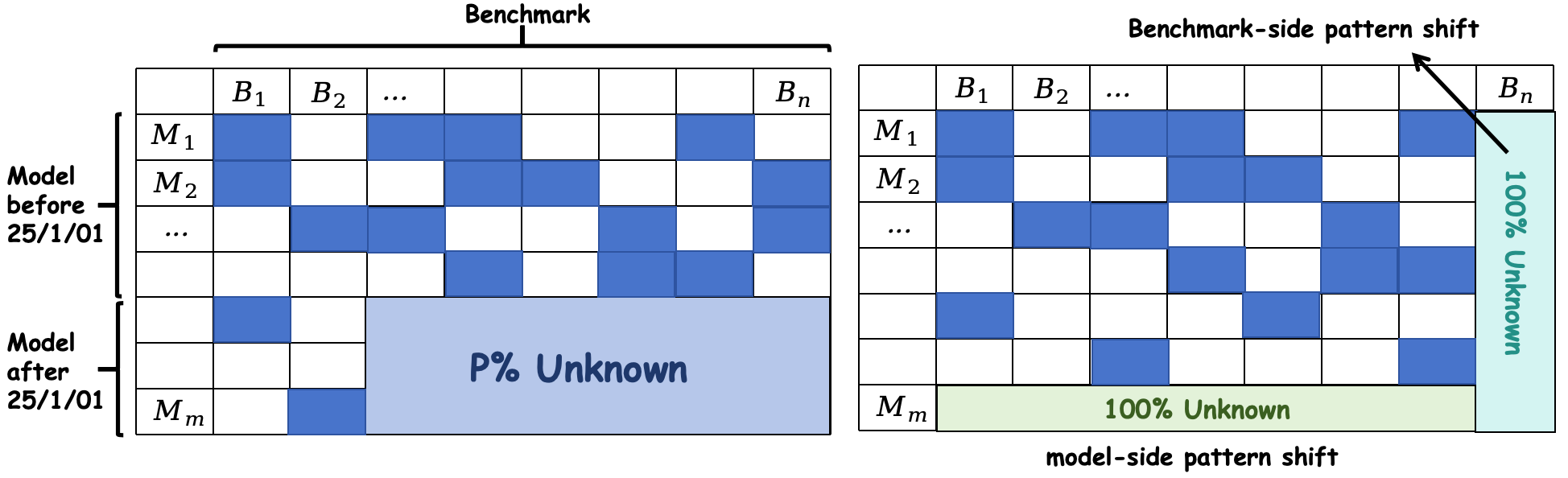}
    \caption{Illustration of evaluation settings. \textbf{Left}: Temporal split for sparsity experiments. Models released before January 2025 have fully observed scores (dark blue), while $P$\% of scores for newer models are masked as prediction targets (light blue). \textbf{Right}: Pattern shift evaluation. Held-out models (bottom row, green) or benchmarks (rightmost column, cyan) are completely unseen during training, testing generalization to novel model characteristics or task categories.}
    \label{fig:split}
\end{figure*}

For newly released models, only a handful of benchmark results may be available. We adopt a temporal split as shown in Figure~\ref{fig:split} (left): 194 models released before January 2025 serve as historical references, while 91 newer models are used for evaluation with $P$\% of their scores randomly masked as prediction targets. This cutoff ensures no knowledge leakage, as our LLM backbone's training data predates January 2025.
We evaluate across sparsity levels $P \in \{40, 60, 80, 95\}$, detailed split statistics are provided in Appendix~\ref{app:aplit}.

\textbf{STAR delivers the most accurate and robust predictions under extreme sparsity.} Table~\ref{tab:model_comparison_95} presents the results under 95\% masking ratio with only 1.4 benchmarks observed per test model on average. We highlight several findings: 
(1) STAR achieves the best performance across all metrics, with RMSE of 8.75, SRCC of 96.10\%, and Total Score of 67.10. The improvements are consistent across both score-based and rank-based metrics, indicating that STAR not only reduces absolute prediction error but also preserves the relative ranking of models more faithfully. 
(2) Compared to the strongest statistical baseline PMF~\cite{zhao2024can}, STAR reduces Score Avg by 31.24\%, improves Rank Avg by 7.52\%, and increases Total Score by 14.46\%. 
(3) Pure LLM and LLM-RAG perform poorly, both underperforming statistical baselines. This demonstrates that LLM reasoning alone is unreliable, and naive retrieval augmentation cannot replace principled statistical modeling.
(4) Semantic-Augmented CPMF already outperforms all baselines whith a total acore of 64.05, and the EVT reasoning module provides further gains of 3.05, validating the effectiveness of our two-stage design: retrieval-augmented statistical expectation followed by evidence-based adjustment.

\noindent \textbf{STAR maintains consistent superiority as sparsity increases.}
Figure~\ref{fig:masking rate} shows performance trends across sparsity levels. Four key observations emerge: (1) STAR consistently achieves the best performance, with its advantage over PMF widening from 2.46 to 8.48 in Total Score as sparsity increases from 40\% to 95\%; (2) Statistical methods like PMF and PCA degrade significantly under high sparsity with decreases of 22.93 and 24.94, while STAR maintains robust performance with only 16.91 decrease; (3) Pure LLM and LLM-RAG remain consistently poor across all settings, confirming that naive LLM-based approaches without statistical grounding are unreliable; (4) Both Semantic-Augmented CPMF and STAR consistently outperform all baselines, validating the effectiveness of integrating semantic features. 
Moreover, the EVT reasoning module provides increasing gains as sparsity grows, from 1.00 at 40\% masking to 3.05 at 95\% masking—demonstrating that evidence-based adjustment becomes more critical when statistical observations are scarce.


\begin{table*}[t]
\centering
\small
\setlength{\tabcolsep}{6pt}
\begin{tabular}{lcccccccccccc}
\toprule
\multirow{2}{*}{\textbf{Scenario}} & \multicolumn{4}{c}{\textbf{Score Avg$\downarrow$}} & \multicolumn{4}{c}{\textbf{Rank Avg(\%)$\uparrow$}} & \multicolumn{4}{c}{\textbf{Total$\uparrow$}} \\
\cmidrule(lr){2-5} \cmidrule(lr){6-9} \cmidrule(lr){10-13}
& PMF & CPMF & STAR & $\Delta$ & PMF & CPMF & STAR & $\Delta$ & PMF & CPMF & STAR & $\Delta$ \\
\midrule
\rowcolor{gray!15}\multicolumn{13}{l}{\textit{Model-side Pattern Shifts}} \\
Architecture & \underline{9.81} & 10.64 & \textbf{9.19} & \textbf{+0.62} & \underline{83.09} & 79.82 & \textbf{83.88} & \textbf{+0.79} & \underline{73.28} & 69.18 & \textbf{74.69} & \textbf{+1.41} \\
Paradigm     & 11.74 & \underline{9.04} & \textbf{8.44} & \textbf{+3.30} & 64.02 & \underline{70.50} & \textbf{71.97} & \textbf{+7.95} & 52.28 & \underline{61.46} & \textbf{63.53} & \textbf{+11.25} \\
Frontier     & 13.22 & \underline{9.82} & \textbf{8.75} & \textbf{+4.47} & 75.53 & \underline{77.32} & \textbf{79.45} & \textbf{+3.92} & 62.31 & \underline{67.50} & \textbf{70.70} & \textbf{+8.39} \\
\textbf{\textit{Average}} & 11.59 & \underline{9.83} & \textbf{8.79} & \textbf{+2.80} & 74.21 & \underline{75.88} & \textbf{78.43} & \textbf{+4.22} & 62.62 & \underline{66.05} & \textbf{69.64} & \textbf{+7.02} \\
\midrule
\rowcolor{gray!15}\multicolumn{13}{l}{\textit{Benchmark-side Pattern Shifts}} \\
Math    & \underline{42.08} & 46.59 & \textbf{13.01} & \textbf{+29.07} & \underline{40.36} & 36.45 & \textbf{60.66} & \textbf{+20.30} & \underline{-1.72} & -10.14 & \textbf{47.65} & \textbf{+49.37} \\
OCR     & 47.60 & \underline{41.70} & \textbf{22.29} & \textbf{+25.31} & 38.82 & \underline{49.48} & \textbf{51.81} & \textbf{+12.99} & -8.78 & \underline{7.78} & \textbf{29.52} & \textbf{+38.30} \\
Chinese & 42.16 & \underline{39.36} & \textbf{14.51} & \textbf{+27.65} & 35.39 & \underline{50.04} & \textbf{51.03} & \textbf{+15.64} & -6.77 & \underline{10.68} & \textbf{36.52} & \textbf{+43.29} \\
\textbf{\textit{Average}} & 43.95 & \underline{42.55} & \textbf{16.60} & \textbf{+27.35} & 38.19 & \underline{45.32} & \textbf{54.50} & \textbf{+16.31} & -5.76 & \underline{2.77} & \textbf{37.90} & \textbf{+43.66} \\
\bottomrule
\end{tabular}
\caption{Performance under pattern shift scenarios. $\Delta$ denotes absolute improvement (STAR $-$ PMF for Rank Avg and Total; PMF $-$ STAR for Score Avg). \textbf{Bold} indicates best; \underline{underline} indicates second best. Details in Appendix \ref{app:patternshift_result}}
\label{tab:table_pattern_shift}
\end{table*}

\subsubsection{Generalization to Pattern Shifts}

Beyond temporal evaluation, we assess generalization to pattern shifts through two scenarios as Figure~\ref{fig:split} (right). For \textbf{model-side shifts}, we hold out models with novel architectures such as MoE, training paradigms like RLHF alignment, or frontier capabilities representing next-generation performance, all in complete cold-start with no known scores. For \textbf{benchmark-side shifts}, we entirely mask specific capability dimensions including Math, OCR, and Chinese during training, requiring cross-benchmark knowledge transfer. Remaining in-distribution entries retain 40\% masking for train-test balance. Split details are in Appendix~\ref{app:patternshift_split}.

\textbf{STAR demonstrates strong generalization under both model-side and benchmark-side pattern shifts}, as shown in Table~\ref{tab:table_pattern_shift}.
A key finding is the stark contrast between the two shift types. Under model-side shifts, improvements are moderate, STAR achieves an average total gain of 7.02 over PMF, suggesting that existing methods already capture known model characteristics reasonably well. Notably, in the architecture scenario, CPMF performs worse than PMF at 69.18 versus 73.28, indicating that semantic features learned from dense models can introduce misleading signals for sparse-activation MoE architectures. In contrast, benchmark-side shifts yield dramatically larger gains. STAR improves the total score by 43.66 on average, indicating that its core strength lies in cross-benchmark generalization.
The benchmark-side results further expose a critical limitation of statistical baselines. PMF produces negative total scores across all three scenarios, while CPMF shows only modest improvement and remains inadequate, with an average total of just 2.77. Such statistical methods fail to transfer effectively to unseen benchmarks.
STAR addresses this limitation through retrieval-augmented reasoning, leveraging semantic similarity to identify relevant references for novel benchmarks. In the Math scenario, STAR reduces the Score Avg from 42.08 to 13.01 while improving Rank Avg from 40.36\% to 60.66\%.

\begin{figure}[h]
    \centering
    \includegraphics[width=\linewidth]{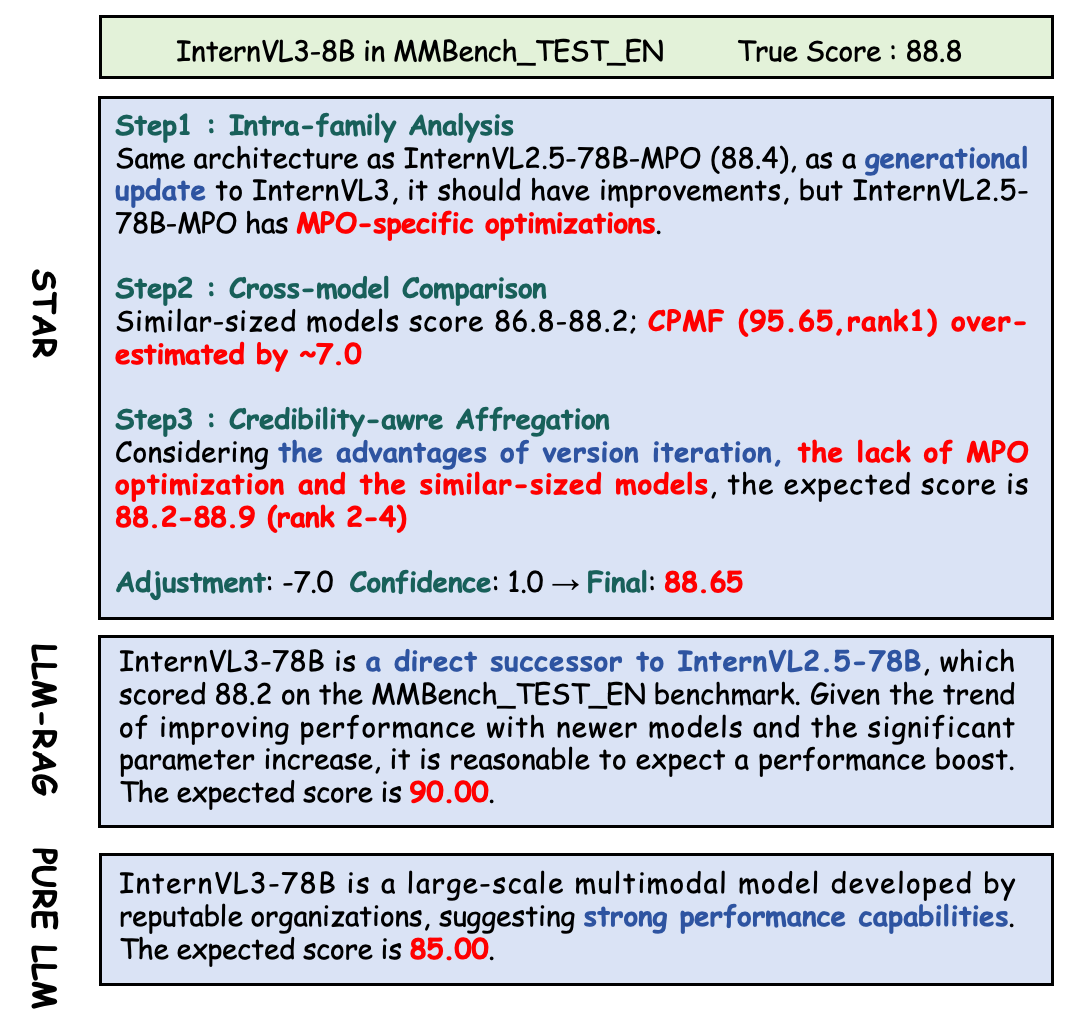}
    \caption{Comparison of reasoning processes for predicting InternVL3-78B on MMBench\_TEST\_EN. Key evidence is highlighted in \textcolor{blue}{blue} for supporting factors and \textcolor{red}{red} for limiting factors.}
    \label{fig:case_study}
\end{figure}
\subsubsection{Explainability}

\textbf{STAR detects and corrects statistical biases through semantic reasoning, producing accurate and traceable predictions} as shown in Figure~\ref{fig:case_study}. Pure LLM~\cite{park2025look} relies on vague descriptions without concrete evidence, resulting in the largest error of 3.8. LLM-RAG identifies the model lineage but assumes newer models always improve, leading to overestimation with error of 1.2. In contrast, STAR follows a structured EVT-guided reasoning process: (1) intra-family analysis identifies InternVL2.5-78B-MPO scoring 88.4 as a reference while noting the absence of MPO-specific optimizations; (2) cross-model comparison reveals that CPMF's prediction of 95.65 is overestimated by 7 points relative to similar-sized models scoring 86.8--88.2; (3) credibility-aware aggregation synthesizes these signals to produce a calibrated prediction of 88.65 with error of only 0.15. Additional success and failure cases are provided in Appendix~\ref{app:cases}.

\subsection{Ablation Experiment}
We conduct three groups of ablation experiments under the 95\% masking rate to analyze how STAR’s design choices contribute to performance. First, we perform component-wise ablations (Tab.\ref{tab:ablation}) to quantify the impact of each module. Second, we examine different LLM backbones (Tab.\ref{tab:llm_backbone}) to assess STAR’s dependence on model capacity and reasoning ability. Third, we evaluate the contribution of each retrieval source (Tab.~\ref{tab:retrieval_sources}), showing how model cards, papers, and web content complement each other.
\begin{table}[h]
\centering
\setlength{\tabcolsep}{3pt}
\small
\begin{tabular}{lccc}
\toprule
\textbf{Method} & \textbf{Score Avg$\downarrow$} & \textbf{Rank Avg$\uparrow$} & \textbf{Total$\uparrow$} \\
\midrule
\rowcolor{gray!15}\multicolumn{4}{l}{\textit{Ablation on Statistical Components}} \\
PMF (w/o features) & 9.00 & 71.79 & 62.79 \\
\quad + Model Features & 8.50 & 72.00 & 63.50 \\
\quad + Benchmark Features & 8.21 & 72.26 & 64.05 \\
\midrule
\rowcolor{gray!15}\multicolumn{4}{l}{\textit{Ablation on Semantic Adjustment}} \\
CPMF + LLM Direct & 8.00 & 72.26 & 64.26 \\
\quad + Family Evolution & 7.77 & 72.82 & 65.05 \\
\quad + Rank-Similar Models & 7.54 & 73.56 & 66.02 \\
\quad + Reasoning (STAR) & \textbf{7.22} & \textbf{74.32} & \textbf{67.10} \\
\bottomrule
\end{tabular}
\caption{Ablation on STAR components. Details in Appendix \ref{app:ablation_com}.}
\label{tab:ablation}
\end{table}

\textbf{Impact of STAR components.} Table~\ref{tab:ablation} demonstrates that STAR benefits from progressively adding semantic and reasoning components. Incorporating statistical features yields only modest improvements over the PMF baseline, increasing the total score by 1.26, from 62.79 to 64.05. In contrast, semantic adjustment modules contribute substantially larger gains, boosting the total score by 2.84, from 64.26 to 67.10, with Rank Avg improving notably from 72.26\% to 74.32\%. Notably, the final reasoning stage delivers the largest single-step improvement, raising the total score from 66.02 to 67.10, which highlights the critical role of explicit evidence-based inference under heavy masking.
\begin{table}[h]
\centering
\setlength{\tabcolsep}{6pt}
\small
\begin{tabular}{lccc}
\toprule
\textbf{LLM Backbone} & \textbf{Score Avg$\downarrow$} & \textbf{Rank Avg$\uparrow$} & \textbf{Total$\uparrow$} \\
\midrule
\rowcolor{gray!15}\multicolumn{4}{l}{\textit{Closed-Source Models}} \\
GPT-5.1 & \textbf{7.22} & \textbf{74.32} & \textbf{67.10} \\
Claude-3.5-Sonnet & 7.28 & 74.18 & 66.90 \\
Gemini-2.5-Pro & 7.33 & 74.05 & 66.72 \\
\midrule
\rowcolor{gray!15}\multicolumn{4}{l}{\textit{Open-Source Models}} \\
Qwen-2.5-72B & 7.68 & 73.24 & 65.56 \\
Qwen-2.5-7B & 8.05 & 72.58 & 64.53 \\
\bottomrule
\end{tabular}
\caption{Impact of LLM backbone on STAR performance. Details in Appendix \ref{app:ablation_backbone}.}
\label{tab:llm_backbone}
\end{table}

\textbf{Impact of LLM backbones.} Table~\ref{tab:llm_backbone} shows that STAR remains highly robust across different LLM backbones. Closed-source frontier models achieve very consistent performance, with Total scores ranging narrowly from 66.72 to 67.10, indicating limited sensitivity to the specific backbone. Open-source models yield slightly lower results, but Qwen-2.5-72B still reaches a competitive Total of 65.56, only 1.54 behind GPT-5.1. Notably, even the lightweight Qwen-2.5-7B achieves a Total score of 64.53, demonstrating that STAR’s effectiveness primarily comes from its retrieval-augmented reasoning framework rather than relying solely on the strongest LLM capacity.

\begin{table}[h]
\centering
\small
\setlength{\tabcolsep}{6pt}
\begin{tabular}{lccc}
\toprule
\textbf{Retrieval Sources} & \textbf{Score Avg$\downarrow$} & \textbf{Rank Avg$\uparrow$} & \textbf{Total$\uparrow$} \\
\midrule

All Sources (Full) & \textbf{7.22} & \textbf{74.32} & \textbf{67.10} \\
\midrule
w/o HuggingFace & 7.57 & 73.46 & 65.89 \\
w/o arXiv & 7.44 & 73.80 & 66.36 \\
w/o Google & 7.35 & 74.04 & 66.69 \\
\midrule
Only HuggingFace & 7.80 & 72.89 & 65.09 \\
Only arXiv & 8.07 & 72.33 & 64.26 \\
Only Google & 8.22 & 72.03 & 63.81 \\
\midrule
No Retrieval 
& 8.64 & 71.16 & 62.52 \\
\bottomrule
\end{tabular}
\caption{Contribution of different retrieval sources. Details in Appendix \ref{app:ablation_retrival}.}
\label{tab:retrieval_sources}
\end{table}
\textbf{Contribution of retrieval sources.} Table~\ref{tab:retrieval_sources} shows that combining all retrieval sources yields the best performance, with STAR benefiting from both structured documentation and complementary web evidence.
We observe: (1) Retrieval is essential for LLM-based adjustment; without external retrieval, the total score drops to 62.52, worse than the CPMF baseline of 64.05. (2) HuggingFace provides the most valuable information: removing it causes the largest drop, from 67.10 to 65.89, as it contains model cards with architecture details, training configurations, and community evaluations essential for cross-model comparison. (3) Sources are complementary, since using a single source yields substantially worse results than the full combination, exemplified by Google alone at 63.81 versus the full score of 67.10, indicating non-redundant contributions: HuggingFace for technical specifications, arXiv for methodology insights, and Google for broader community feedback.

\section{Conclusion}
We presented STAR, a framework that bridges statistical expectations and agentic reasoning for large model benchmark prediction. Drawing on Expectation Violation Theory (EVT) from cognitive science, STAR first derives statistical expectations through retrieval-augmented PMF, then refines them via structured evidence analysis, including intra-family evolution, cross-model comparison, and credibility-aware aggregation. Experiments show that STAR reduces prediction error by 31.24\% and improves rank consistency by 7.52\% over the strongest baseline, while providing traceable explanations. We believe this work offers a promising step toward integrating data-driven and knowledge-driven approaches for robust and interpretable model evaluation.

\section*{Impact Statement}

In this work, we propose STAR, a sample-efficient framework for predicting benchmark performance of large models. On the positive side, our method can help community developers and practitioners quickly perform model selection, reducing evaluation workload and allowing them to screen for models that better match their specific needs under limited compute budgets. Our EVT-guided reasoning framework may also inspire other agent-based systems and be applied to a broader range of performance prediction tasks beyond standardized benchmarks. On the negative side, there is a risk that users may over-rely on predicted scores instead of conducting thorough empirical and safety evaluations. Overall, we believe STAR is a powerful and practical tool to make model evaluation more efficient and accessible, as long as it is used as a complement rather than a replacement for careful testing and governance.


\bibliography{example_paper}
\bibliographystyle{icml2026}

\newpage
\appendix
\onecolumn
\section{Dataset Details}
\label{app:dataset}

\subsection{Benchmark Descriptions}
\label{app:benchmark_description}

Table~\ref{tab:app_benchmarks} provides detailed information about the 28 benchmarks used in our evaluation, spanning 11 categories: General (10), Knowledge (1), Culture (1), Math (1), Science (3), Reasoning (3), Perception (1), Hallucination (2), OCR (3), Multilingual (1), and Quality (2). These benchmarks collectively cover diverse VLM capabilities including multimodal understanding, domain-specific reasoning, visual perception, hallucination detection, and text-rich image comprehension. Dataset sizes range from 60 (LLaVABench) to 31,325 (MMT-Bench\_VAL) samples, with most using accuracy as the evaluation metric.

\begin{table*}[htbp]
\centering
\caption{Complete list of 28 benchmarks in our VLM evaluation dataset.}
\label{tab:app_benchmarks}
\small
\begin{tabular}{llccp{7.5cm}}
\toprule
\textbf{Benchmark} & \textbf{Category} & \textbf{\#Samples} & \textbf{Metric} & \textbf{Description} \\
\midrule
SEEDBench\_IMG & General & 19,000 & Acc & Image understanding \\
MMBench\_TEST\_EN & General & 3,000+ & Acc & Comprehensive multimodal understanding (English) \\
MMBench\_TEST\_CN & General & 3,000+ & Acc & Comprehensive multimodal understanding (Chinese) \\
MMBench\_TEST\_EN\_V11 & General & 3,000+ & Acc & Updated MMBench with refined evaluation (English) \\
MMBench\_TEST\_CN\_V11 & General & 3,000+ & Acc & Updated MMBench with refined evaluation (Chinese) \\
MME & General & 2,374 & Score & Perception and cognition across 14 subtasks \\
MMVet & General & 218 & Score & Complex tasks integrating multiple VL capabilities \\
LLaVABench & General & 60 & Score & Multimodal instruction following evaluation \\
SEEDBench2 & General & 24,000 & Acc & Interleaved multimodal understanding \\
SEEDBench2\_Plus & General & 24,000 & Acc & Extended SEED with more evaluation dimensions \\
\midrule
MMMU\_VAL & Knowledge & 900 & Acc & College-level multi-discipline reasoning \\
CCBench & Culture & 510 & Acc & Chinese cultural knowledge understanding \\
\midrule
MathVista & Math & 6,141 & Acc & Mathematical reasoning in visual contexts \\
AI2D & Science & 5,000 & Acc & Science diagram interpretation and reasoning \\
ScienceQA\_VAL & Science & 4,241 & Acc & Multimodal science question answering (Val) \\
ScienceQA\_TEST & Science & 4,241 & Acc & Multimodal science question answering (Test) \\
\midrule
MMStar & Reasoning & 1,500 & Acc & Vision-dependent multi-modal reasoning \\
RealWorldQA & Reasoning & 700 & Acc & Real-world spatial reasoning from vehicle images \\
BLINK & Perception & 3,807 & Acc & Visual perception across 14 classic CV tasks \\
MMT-Bench\_VAL & Reasoning & 31,325 & Acc & Expert knowledge visual recognition and planning \\
\midrule
HallusionBench & Hallucination & 1,129 & Acc & Visual illusion and hallucination detection \\
POPE & Hallucination & 3,000 & Acc & Object hallucination evaluation \\
\midrule
TextVQA\_VAL & OCR & 5,000 & Acc & Reading text in images to answer questions \\
ChartQA\_TEST & Chart & 23,100 & Acc & Chart reasoning and question answering \\
OCRVQA\_TESTCORE & OCR & 500 & Acc & OCR-based VQA on book covers \\
MTVQA & Multilingual & 6,778 & Acc & Multilingual text understanding in images \\
\midrule
QBench & Quality & 3,499 & Acc & Low-level image quality perception \\
ABench & Quality & 2,864 & Acc & AI-generated image quality assessment \\
\bottomrule
\end{tabular}
\end{table*}

\subsection{Evaluation Metrics}
\label{sec:metrics}
We evaluate performance prediction using eight metrics in two categories. All scores are normalized to the 0--100 range before computing metrics.

\subsubsection{Score-based Metrics ($\downarrow$)}

\textbf{Mean Absolute Error (MAE)} measures the average absolute difference between predicted and true scores, providing a direct interpretation of prediction error magnitude.
\begin{equation}
\text{MAE} = \frac{1}{N} \sum_{i=1}^{N} \left| \hat{y}_i - y_i \right|
\end{equation}

\textbf{Root Mean Square Error (RMSE)} also measures prediction error but penalizes larger errors more heavily, making it more sensitive to outliers.
\begin{equation}
\text{RMSE} = \sqrt{\frac{1}{N} \sum_{i=1}^{N} \left( \hat{y}_i - y_i \right)^2}
\end{equation}
where $\hat{y}_i$ and $y_i$ denote predicted and ground-truth scores, and $N$ is the number of test samples.

\textbf{Score Average} combines both error metrics to provide a unified measure of score-level prediction accuracy.
\begin{equation}
\text{Score}_{\text{avg}} = \frac{\text{MAE} + \text{RMSE}}{2}
\end{equation}

\subsubsection{Rank-based Metrics ($\uparrow$)}

All rank-based metrics are normalized to $[0\%, 100\%]$, where higher values indicate better ranking preservation.

\textbf{Spearman's Rank Correlation Coefficient (SRCC)} measures how well predicted scores preserve the true ranking order based on rank differences.
\begin{equation}
\text{SRCC} = \frac{1 + \rho}{2} \times 100\%, \quad \rho = 1 - \frac{6 \sum_{i=1}^{N} d_i^2}{N(N^2 - 1)}
\end{equation}
where $d_i$ is the difference between predicted and true ranks of sample $i$.

\textbf{Kendall's Rank Correlation Coefficient (KRCC)} assesses ranking consistency by counting concordant and discordant pairs, and is more robust to outliers than SRCC.
\begin{equation}
\text{KRCC} = \frac{1 + \tau}{2} \times 100\%, \quad \tau = \frac{C - D}{\frac{1}{2} N(N-1)}
\end{equation}
where $C$ and $D$ are the numbers of concordant and discordant pairs.

\textbf{MAE@$k$} measures the proportion of predictions whose rank error is within $k$ positions, directly reflecting practical ranking accuracy.
\begin{equation}
\text{MAE@}k = \frac{100\%}{N} \sum_{i=1}^{N} \mathbb{1}\left[ \left| r_i^{(\text{pred})} - r_i^{(\text{true})} \right| \leq k \right]
\end{equation}
We use $k=3$ in our experiments.

\textbf{Rank Average} aggregates all rank-based metrics to provide a unified measure of ranking quality.
\begin{equation}
\text{Rank}_{\text{avg}} = \frac{\text{SRCC} + \text{KRCC} + \text{MAE@3}}{3}
\end{equation}

\subsubsection{Total Score ($\uparrow$)}

\textbf{Total Score} serves as a unified indicator that balances both prediction accuracy and ranking quality.
\begin{equation}
\text{Total Score} = \text{Rank}_{\text{avg}} - \text{Score}_{\text{avg}}
\end{equation}
Since Rank Avg (\%) and Score Avg are both on comparable 0--100 scales, this difference directly measures the net gain of ranking quality over prediction error. Higher values indicate better overall performance.
\section{Supplementary Results For Sparsity Performance}
\subsection{Train/Test Split Statistics}
\label{app:aplit}

Table~\ref{tab:app_split_stats} summarizes the statistics of our temporal data split under different masking ratios. Our dataset contains 195 models released before January 2025 and 91 models released afterward. The prediction targets are the masked scores of the 91 newer models. For training, we utilize two sources of observed data: (1) all available scores from the 194 older models, and (2) the unmasked scores from the 91 test models, reflecting the realistic scenario where some benchmark results become available shortly after model release.
As the masking ratio increases, fewer benchmark scores remain observable for test models: at 40\% masking, each test model retains an average of 16.8 out of 28 benchmarks for training, whereas at 95\% masking, only 1.4 benchmarks are available on average. 

\begin{table*}[htbp]
\centering
\caption{Dataset split statistics under different masking ratios. 
}
\label{tab:app_split_stats}
\begin{tabular}{lcccc}
\toprule
& \textbf{40\% mask} & \textbf{60\% mask} & \textbf{80\% mask} & \textbf{95\% mask} \\
\midrule
Train \#Entries & 5,411 & 5,113 & 4,761 & 4,544 \\
Test \#Entries & 621 & 919 & 1,271 & 1,488 \\
Avg. Benchmark & 16.8/28 & 11.2/28 & 5.6/28 & 1.4/28 \\
\bottomrule
\end{tabular}
\end{table*}

\subsection{Results Across Different Sparsity Levels}
\label{app:sparsity_result}

We provide detailed performance comparisons under 40\%, 60\%, and 80\% masking ratios in Tables~\ref{tab:app_model_comparison_40}, \ref{tab:app_model_comparison_60}, and \ref{tab:app_model_comparison_80}, respectively. STAR consistently achieves the best performance across all metrics and sparsity levels, followed by Semantic-Augmented CPMF as the second best.

\textbf{Analysis of 40\% Masking.}
Under 40\% masking, each test model observes approximately 16.8 benchmarks on average. In this data-rich regime, STAR achieves a Total Score of 84.01, followed by CPMF at 83.01 and PMF at 81.55. STAR outperforms PMF by 2.46 points, representing a 3.0\% relative improvement. PMF performs remarkably well with RMSE of 7.17 and SRCC of 96.89\%, demonstrating that matrix factorization is highly effective when sufficient observations are available. Pure LLM achieves the lowest Total Score of 34.73, confirming that LLM reasoning without statistical grounding suffers from hallucination regardless of data availability. The relatively small gap between STAR and statistical baselines suggests that semantic reasoning provides incremental gains when statistical signals are strong.

\textbf{Analysis of 60\% Masking.}
At 60\% masking, each test model observes approximately 11.2 benchmarks. STAR maintains its lead with a Total Score of 78.62, while PMF drops to 75.30. The gap between STAR and PMF widens to 3.32, corresponding to a 4.4\% relative improvement. This indicates that semantic augmentation becomes increasingly valuable as observations decrease. PCA experiences the steepest decline, dropping from 71.32 to 60.83 with a 14.7\% decrease, while STAR shows the most stable degradation from 84.01 to 78.62 with only a 6.4\% decrease. Pure LLM and LLM-RAG continue to underperform with Total Scores of 34.45 and 50.69 respectively, confirming that naive LLM-based approaches cannot substitute for principled statistical modeling.

\textbf{Analysis of 80\% Masking.}
Under 80\% masking, each test model observes only approximately 5.6 benchmarks. STAR achieves a Total Score of 72.97, outperforming CPMF at 71.75 and PMF at 67.48. The gap between STAR and PMF reaches 5.49, representing an 8.1\% relative improvement. This demonstrates that semantic reasoning becomes critical under extreme sparsity. Statistical methods show substantial degradation: PMF drops 17.2\% compared to 40\% masking, while PCA collapses by 29.2\%. The variance also increases notably. PMF grows from $\pm$0.38 to $\pm$0.58, while STAR increases more moderately from $\pm$0.52 to $\pm$0.72, indicating more stable predictions. The gap between STAR and CPMF widens from 1.0 at 40\% masking to 1.22 at 80\% masking, validating that evidence-based reasoning contributes more when statistical signals are weak.

\begin{table*}[h]
\centering
\small
\setlength{\tabcolsep}{4.5pt}
\begin{tabular}{lcccccccc}
\toprule
 \multirow{2}{*}{\textbf{Method}} & \multicolumn{3}{c}{\textbf{Score-based Metrics$\downarrow$}} & \multicolumn{4}{c}{\textbf{Rank-based Metrics$\uparrow$}} & \multicolumn{1}{c}{\textbf{Total$\uparrow$}} \\
\cmidrule(lr){2-4} \cmidrule(lr){5-8} \cmidrule(lr){9-9}
& RMSE & MAE & Score Avg & SRCC(\%) & KRCC(\%) & MAE@3(\%) & Rank Avg(\%) & Score \\
\midrule
\rowcolor{gray!15}\multicolumn{9}{l}{\textit{Statistical Baselines}} \\
 
Global Mean & 16.03 & 12.51 & 14.27 & 83.79 & 75.46 & 23.51 & 60.92 & 46.65 \\
Mean Of Means & 12.87 & 9.46 & 11.17 & 91.83 & 83.65 & 74.24 & 83.24 & 72.07 \\
\midrule
\rowcolor{gray!15}\multicolumn{9}{l}{\textit{Statistic-based Methods}} \\

PCA~\cite{ruan2024observational} & 11.53 & 7.28 & 9.41 & 93.22 & 85.19 & 63.77 & 80.73 & 71.32 {\scriptsize $\pm$ 0.45} \\
Regression~\cite{polo2024sloth} & 11.66 & 7.28 & 9.47 & 92.60 & 85.32 & 67.55 & 81.82 & 72.35 {\scriptsize $\pm$ 0.42} \\
NCF~\cite{zhang2024collaborative} & 13.58 & 8.91 & 11.25 & 90.17 & 81.61 & 46.38 & 72.72 & 61.47 {\scriptsize $\pm$ 0.55} \\
PMF~\cite{zhao2024can} & 7.17 & 5.03 & 6.10 & 96.89 & 89.89 & 76.17 & 87.65 & 81.55 {\scriptsize $\pm$ 0.38} \\
\midrule
\rowcolor{gray!15}\multicolumn{9}{l}{\textit{LLM-based Methods}} \\

Pure LLM~\cite{park2025look} & 23.60 & 16.43 & 20.02 & 72.47 & 66.88 & 24.89 & 54.75 & 34.73 {\scriptsize $\pm$ 0.95} \\
LLM-RAG & 13.93 & 7.93 & 10.93 & 81.71 & 66.90 & 53.62 & 67.41 & 56.48 {\scriptsize $\pm$ 0.82} \\
\midrule

Semantic-Augmented CPMF & \underline{6.38} & \underline{4.33} & \underline{5.36} & \underline{97.47} & \underline{91.27} & \underline{76.38} & \underline{88.37} & \underline{83.01} {\scriptsize $\pm$ 0.42} \\

STAR & \textbf{6.13} & \textbf{3.98} & \textbf{5.06} & \textbf{97.65} & \textbf{91.73} & \textbf{77.83} & \textbf{89.07} & \textbf{84.01} {\scriptsize $\pm$ 0.52} \\

\bottomrule
\end{tabular}
\caption{Performance comparison under 40\% masking. Results are reported as mean {\scriptsize $\pm$ std} over 5 runs.}
\label{tab:app_model_comparison_40}
\end{table*}

\begin{table*}[h]
\centering
\small
\setlength{\tabcolsep}{4.5pt}
\begin{tabular}{lcccccccc}
\toprule
 \multirow{2}{*}{\textbf{Method}} & \multicolumn{3}{c}{\textbf{Score-based Metrics$\downarrow$}} & \multicolumn{4}{c}{\textbf{Rank-based Metrics$\uparrow$}} & \multicolumn{1}{c}{\textbf{Total$\uparrow$}} \\
\cmidrule(lr){2-4} \cmidrule(lr){5-8} \cmidrule(lr){9-9}
& RMSE & MAE & Score Avg & SRCC(\%) & KRCC(\%) & MAE@3(\%) & Rank Avg(\%) & Score \\
\midrule
\rowcolor{gray!15}\multicolumn{9}{l}{\textit{Statistical Baselines}} \\

Global Mean & 16.14 & 12.68 & 14.41 & 85.83 & 77.31 & 16.76 & 59.97 & 45.56 \\

Mean Of Means & 12.93 & 9.50 & 11.21 & 92.82 & 84.81 & 60.61 & 79.41 & 68.20 \\

\midrule
\rowcolor{gray!15}\multicolumn{9}{l}{\textit{Statistic-based Methods}} \\

PCA~\cite{ruan2024observational} & 12.83 & 8.86 & 10.84 & 91.43 & 83.10 & 40.48 & 71.67 & 60.83 {\scriptsize $\pm$ 0.58} \\

Regression~\cite{polo2024sloth} & 11.60 & 7.59 & 9.60 & 93.36 & 85.78 & 56.61 & 78.58 & 68.98 {\scriptsize $\pm$ 0.52} \\

NCF~\cite{zhang2024collaborative} & 13.03 & 8.72 & 10.88 & 91.25 & 82.69 & 31.34 & 68.43 & 57.55 {\scriptsize $\pm$ 0.65} \\

PMF~\cite{zhao2024can} & 9.35 & 5.33 & 7.34 & 96.34 & 89.88 & 61.70 & 82.64 & 75.30 {\scriptsize $\pm$ 0.48} \\

\midrule
\rowcolor{gray!15}\multicolumn{9}{l}{\textit{LLM-based Methods}} \\

Pure LLM~\cite{park2025look} & 23.55 & 16.72 & 20.13 & 72.94 & 66.92 & 23.88 & 54.58 & 34.45 {\scriptsize $\pm$ 1.05} \\
LLM-RAG & 14.46 & 8.55 & 11.50 & 80.45 & 65.52 & 40.59 & 62.19 & 50.69 {\scriptsize $\pm$ 0.92} \\

\midrule

Semantic-Augmented CPMF & \underline{6.90} & \underline{4.61} & \underline{5.76} & \underline{96.77} & \underline{90.66} & \underline{62.68} & \underline{83.37} & \underline{77.61} {\scriptsize $\pm$ 0.52} \\

STAR & \textbf{6.77} & \textbf{4.34} & \textbf{5.56} & \textbf{97.56} & \textbf{91.48} & \textbf{63.50} & \textbf{84.18} & \textbf{78.62} {\scriptsize $\pm$ 0.62} \\

\bottomrule
\end{tabular}
\caption{Performance comparison under 60\% masking. Results are reported as mean {\scriptsize $\pm$ std} over 5 runs.}
\label{tab:app_model_comparison_60}
\end{table*}

\begin{table*}[t]
\centering
\small
\setlength{\tabcolsep}{4.5pt}
\begin{tabular}{lcccccccc}
\toprule
 \multirow{2}{*}{\textbf{Method}} & \multicolumn{3}{c}{\textbf{Score-based Metrics$\downarrow$}} & \multicolumn{4}{c}{\textbf{Rank-based Metrics$\uparrow$}} & \multicolumn{1}{c}{\textbf{Total$\uparrow$}} \\
\cmidrule(lr){2-4} \cmidrule(lr){5-8} \cmidrule(lr){9-9}
& RMSE & MAE & Score Avg & SRCC(\%) & KRCC(\%) & MAE@3(\%) & Rank Avg(\%) & Score \\
\midrule
\rowcolor{gray!15}\multicolumn{9}{l}{\textit{Statistical Baselines}} \\
 
Global Mean & 16.18 & 12.94 & 14.56 & 85.56 & 76.95 & 12.12 & 58.21 & 43.65 \\
Mean Of Means & 12.88 & 9.69 & 11.29 & 92.73 & 84.56 & 47.36 & 74.89 & 63.60 \\
\midrule
\rowcolor{gray!15}\multicolumn{9}{l}{\textit{Statistic-based Methods}} \\

PCA~\cite{ruan2024observational} & 14.54 & 10.81 & 12.67 & 88.03 & 79.26 & 22.19 & 63.16 & 50.49 {\scriptsize $\pm$ 0.72} \\
Regression~\cite{polo2024sloth} & 11.98 & 8.19 & 10.09 & 92.72 & 84.78 & 44.86 & 74.12 & 64.03 {\scriptsize $\pm$ 0.62} \\
NCF~\cite{zhang2024collaborative} & 11.51 & 7.79 & 9.65 & 92.68 & 84.19 & 28.64 & 68.50 & 58.85 {\scriptsize $\pm$ 0.75} \\

PMF~\cite{zhao2024can} & 10.56 & 6.81 & 8.69 & 95.02 & 86.87 & 46.52 & 76.17 & 67.48 {\scriptsize $\pm$ 0.58} \\
\midrule
\rowcolor{gray!15}\multicolumn{9}{l}{\textit{LLM-based Methods}} \\

Pure LLM~\cite{park2025look} & 23.50 & 16.63 & 20.07 & 72.03 & 66.26 & 24.27 & 54.25 & 34.18 {\scriptsize $\pm$ 1.15} \\
LLM-RAG & 14.72 & 8.98 & 11.85 & 77.50 & 62.29 & 29.35 & 56.38 & 44.53 {\scriptsize $\pm$ 1.02} \\
\midrule

Semantic-Augmented CPMF & \underline{8.75} & \underline{5.36} & \underline{7.06} & \underline{96.50} & \underline{89.64} & \underline{50.28} & \underline{78.81} & \underline{71.75} {\scriptsize $\pm$ 0.62} \\

STAR & \textbf{7.50} & \textbf{4.92} & \textbf{6.21} & \textbf{97.04} & \textbf{90.18} & \textbf{51.31} & \textbf{79.18} & \textbf{72.97} {\scriptsize $\pm$ 0.72} \\

\bottomrule
\end{tabular}
\caption{Performance comparison under 80\% masking. Results are reported as mean {\scriptsize $\pm$ std} over 5 runs.}
\label{tab:app_model_comparison_80}
\end{table*}

\section{Supplementary Results For Generalization To Pattern Shifts}

\subsection{Data Split Details}
\label{app:patternshift_split}

We evaluate generalization under pattern shifts that commonly occur in practice. Table~\ref{tab:app_pattern_shift_data} summarizes the data split statistics for six scenarios categorized into two types.

\textbf{Model-side Pattern Shifts.} These scenarios evaluate prediction accuracy when test models exhibit characteristics unseen during training, simulating real-world situations where newly released models introduce novel architectures or training paradigms. We design three scenarios: (1) \textit{Architecture Shift}: We hold out all 10 mixture-of-experts (MoE) models from training, probing generalization to emerging sparse-activation architectures that differ fundamentally from dense models. (2) \textit{Paradigm Shift}: All 30 RLHF/DPO-aligned models are excluded, evaluating robustness to training methodology variations and simulating breakthroughs in alignment techniques. (3) \textit{Frontier Shift}: We reserve the top-20 highest-performing models for evaluation, mimicking the practical challenge of predicting capabilities of next-generation frontier models before their benchmarks are publicly available. All held-out models are in complete cold-start settings with no historical scores in training, while other models are randomly masked at 40\% to simulate realistic data sparsity.

\textbf{Benchmark-side Pattern Shifts.} These scenarios test generalization to entirely new task categories, representing situations where practitioners need predictions on newly proposed evaluation benchmarks. We fully mask specific benchmark groups during training: MathVista for Math reasoning, TextVQA/ChartQA/OCRVQA for OCR understanding, and CCBench/MMBench\_CN for Chinese-language comprehension. This requires the model to transfer knowledge from semantically related benchmarks to predict performance on these unseen task categories—a particularly challenging setting for statistical methods that have no direct observations to learn benchmark-specific patterns.

\begin{table}[h]
\centering
\caption{Data split statistics for pattern shift experiments. ``\#Train'' and ``\#Test'' denote the number of models in each split. ``\#Train Samples'' and ``\#Test Samples'' indicate the number of non-empty (model, benchmark) pairs available for training and evaluation.}
\label{tab:app_pattern_shift_data}
\begin{tabular}{llccccc}
\toprule
\textbf{Category} & \textbf{Scenario} & \textbf{Held-out Target} & \textbf{\#Train} & \textbf{\#Test} & \textbf{\#Train Samples} & \textbf{\#Test Samples} \\
\midrule
\multirow{3}{*}{Model-side} 
  & Architecture & 10 MoE models       & 275 & 10 & 3,535 & 167 \\
  & Paradigm     & 30 RLHF/DPO models  & 255 & 30 & 3,194 & 762 \\
  & Frontier     & 20 Top models       & 265 & 20 & 3,438 & 314 \\
\midrule
\multirow{3}{*}{Benchmark-side}
  & Math    & 1 benchmark             & 285 & 285 & 3,450 & 284 \\
  & OCR     & 3 benchmarks & 285 & 285 & 3,427 & 352 \\
  & Chinese & 3 benchmarks   & 285 & 285 & 3,180 & 743 \\
\bottomrule
\end{tabular}
\end{table}

\subsection{Results For Generalization To Pattern Shifts}
\label{app:patternshift_result}

Table~\ref{tab:app_gen} presents the generalization results across six pattern shift scenarios. For model-side shifts, STAR consistently achieves the best performance, scoring 74.69, 63.53, and 70.70 on Architecture, Paradigm, and Frontier scenarios respectively. Notably, CPMF underperforms PMF in Architecture shift at 69.18 versus 73.28, as semantic features learned from dense models introduce misleading signals for MoE architectures. For benchmark-side shifts where target benchmarks have no training data, statistical methods completely fail with PMF achieving negative Total Scores of -1.72, -8.78, and -6.77, while STAR dramatically outperforms at 47.65, 29.52, and 36.52 through effective knowledge transfer. The variance analysis further reveals that STAR achieves substantially lower variance in benchmark-side scenarios at $\pm$1.58 to $\pm$1.82 compared to PMF at $\pm$3.42 to $\pm$3.85, indicating more stable predictions when generalizing to unseen task categories.

\begin{table*}[h]
\centering
\caption{Generalization Results. STAR consistently outperforms statistical baselines, with particularly large gains on Benchmark-side shifts where target benchmarks have no training data. Total Score = Rank Avg $-$ Score Avg. Results are reported as mean {\scriptsize $\pm$ std} over 5 runs.}
\label{tab:app_gen}
\small
\begin{tabular}{llcccccccc}
\toprule
\multirow{2}{*}{\textbf{Scenario}} & \multirow{2}{*}{\textbf{Method}} & \multicolumn{3}{c}{\textbf{Score-based}$\downarrow$} & \multicolumn{4}{c}{\textbf{Rank-based}$\uparrow$} & \multirow{2}{*}{\textbf{Total}$\uparrow$} \\
\cmidrule(lr){3-5} \cmidrule(lr){6-9}
& & RMSE & MAE & Score Avg & SRCC\% & KRCC\% & MAE@3\% & Rank Avg\% & \\
\midrule
\multirow{3}{*}{Architecture (MoE)} 
& PMF  & 11.01 & 8.61 & 9.81 & 90.16 & 81.40 & 77.71 & 83.09 & 73.28 {\scriptsize $\pm$ 0.52} \\
& CPMF & 11.73 & 9.55 & 10.64 & 88.08 & 78.50 & 72.89 & 79.82 & 69.18 {\scriptsize $\pm$ 0.58} \\
& STAR & \textbf{10.68} & \textbf{7.69} & \textbf{9.19} & \textbf{89.43} & \textbf{81.50} & \textbf{80.72} & \textbf{83.88} & \textbf{74.69} {\scriptsize $\pm$ 0.68} \\
\midrule
\multirow{3}{*}{Paradigm (RLHF)} 
& PMF  & 13.15 & 10.32 & 11.74 & 88.51 & 79.54 & 24.02 & 64.02 & 52.28 {\scriptsize $\pm$ 0.65} \\
& CPMF & 10.14 & 7.94 & 9.04 & 91.88 & 83.26 & 36.35 & 70.50 & 61.46 {\scriptsize $\pm$ 0.62} \\
& STAR & \textbf{9.55} & \textbf{7.33} & \textbf{8.44} & \textbf{92.48} & \textbf{84.07} & \textbf{39.37} & \textbf{71.97} & \textbf{63.53} {\scriptsize $\pm$ 0.78} \\
\midrule
\multirow{3}{*}{Frontier (Top-20)} 
& PMF  & 14.09 & 12.35 & 13.22 & 92.66 & 83.28 & 50.64 & 75.53 & 62.31 {\scriptsize $\pm$ 0.58} \\
& CPMF & 10.71 & 8.93 & 9.82 & 94.97 & 86.35 & 50.64 & 77.32 & 67.50 {\scriptsize $\pm$ 0.55} \\
& STAR & \textbf{9.71} & \textbf{7.79} & \textbf{8.75} & \textbf{95.55} & \textbf{87.38} & \textbf{55.41} & \textbf{79.45} & \textbf{70.70} {\scriptsize $\pm$ 0.72} \\
\midrule
\multirow{3}{*}{Benchmark (Math)} 
& PMF  & 44.21 & 39.94 & 42.08 & 61.33 & 57.64 & 2.11 & 40.36 & -1.72 {\scriptsize $\pm$ 3.52} \\
& CPMF & 48.63 & 44.55 & 46.59 & 53.89 & 52.63 & 2.82 & 36.45 & -10.14 {\scriptsize $\pm$ 4.25} \\
& STAR & \textbf{14.70} & \textbf{11.31} & \textbf{13.01} & \textbf{86.09} & \textbf{77.18} & \textbf{18.70} & \textbf{60.66} & \textbf{47.65} {\scriptsize $\pm$ 1.68} \\
\midrule
\multirow{3}{*}{Benchmark (OCR)} 
& PMF  & 50.65 & 44.55 & 47.60 & 55.73 & 53.90 & 6.82 & 38.82 & -8.78 {\scriptsize $\pm$ 3.85} \\
& CPMF & 44.95 & 38.44 & 41.70 & 71.76 & 65.61 & 11.08 & 49.48 & 7.78 {\scriptsize $\pm$ 3.28} \\
& STAR & \textbf{25.18} & \textbf{19.41} & \textbf{22.29} & \textbf{70.64} & \textbf{67.62} & \textbf{17.16} & \textbf{51.81} & \textbf{29.52} {\scriptsize $\pm$ 1.82} \\
\midrule
\multirow{3}{*}{Benchmark (Chinese)} 
& PMF  & 45.40 & 38.91 & 42.16 & 52.08 & 51.40 & 2.69 & 35.39 & -6.77 {\scriptsize $\pm$ 3.42} \\
& CPMF & 42.28 & 36.44 & 39.36 & 75.01 & 67.56 & 7.54 & 50.04 & 10.68 {\scriptsize $\pm$ 2.95} \\
& STAR & \textbf{16.94} & \textbf{12.07} & \textbf{14.51} & \textbf{73.70} & \textbf{67.35} & \textbf{12.03} & \textbf{51.03} & \textbf{36.52} {\scriptsize $\pm$ 1.58} \\
\bottomrule
\end{tabular}
\end{table*}

\section{Case Study}
\label{app:cases}

We present case studies to illustrate when STAR's reasoning succeeds and fails, providing insights into the strengths and limitations of our approach.

\begin{figure}[h]
    \centering
    \includegraphics[width=\linewidth]{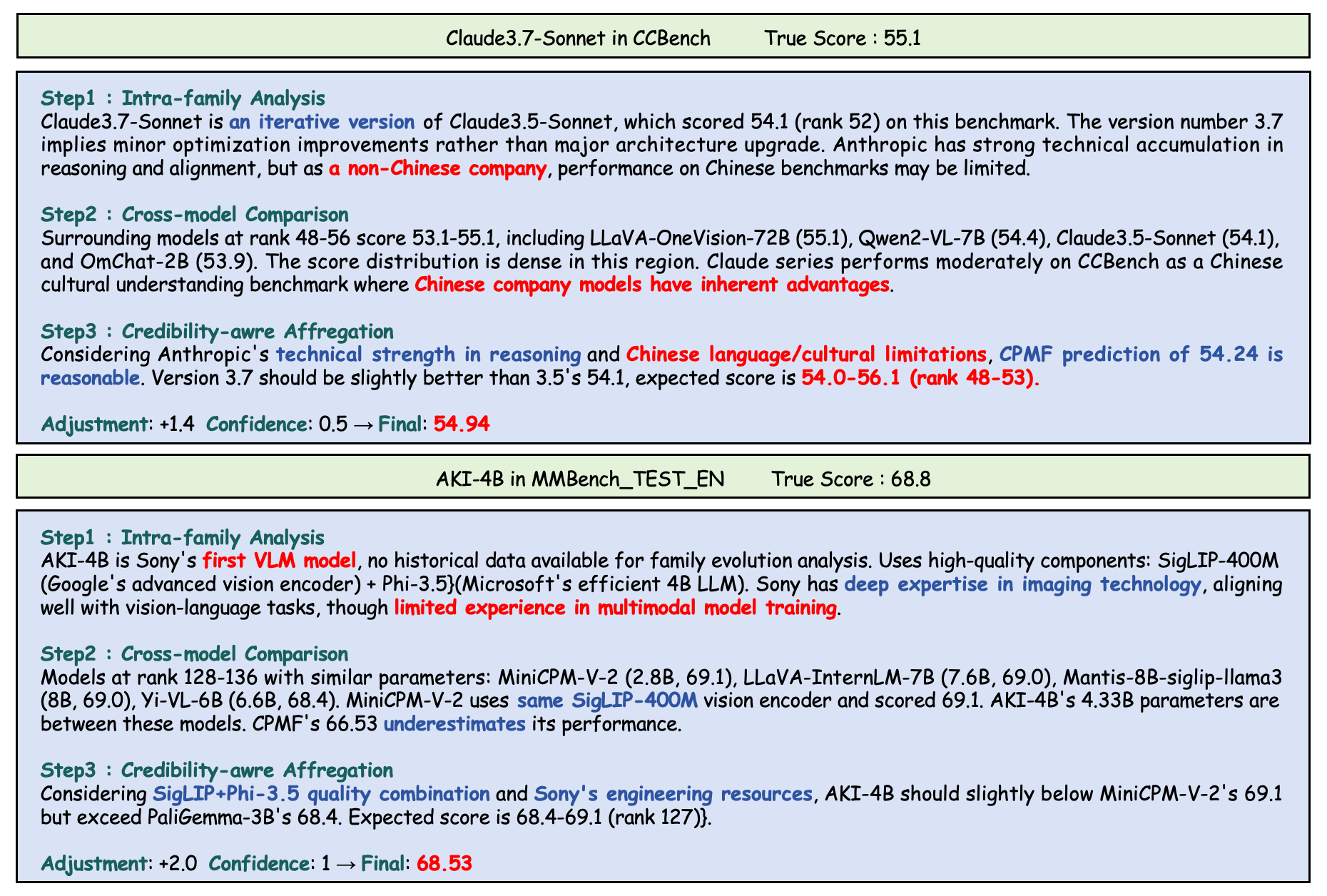}
    \caption{Successful prediction cases. STAR achieves prediction errors of only 0.16 and 0.27 for Claude3.7-Sonnet on CCBench and AKI-4B on MMBench\_TEST\_EN respectively. Key success factors are highlighted: family references in blue, capability indicators in green, and contextual limitations in red.}
    \label{fig:good_case}
\end{figure}

\textbf{Analysis of Successful Cases.}
The first case demonstrates effective intra-family reasoning. STAR correctly identifies Claude3.7-Sonnet as an iterative version of Claude3.5-Sonnet, which scored 54.1 on CCBench. By recognizing that version 3.7 implies minor improvements rather than major upgrades, and accounting for Anthropic's limited advantage on Chinese cultural benchmarks, STAR predicts 54.94 with only 0.16 error. The second case shows successful cross-model comparison for a cold-start model. Despite AKI-4B being Sony's first VLM with no family history, STAR identifies MiniCPM-V-2 as a reliable reference due to sharing the same SigLIP-400M vision encoder, correctly adjusting CPMF's underestimate of 66.53 to 68.53.

\begin{figure}[h]
    \centering
    \includegraphics[width=\linewidth]{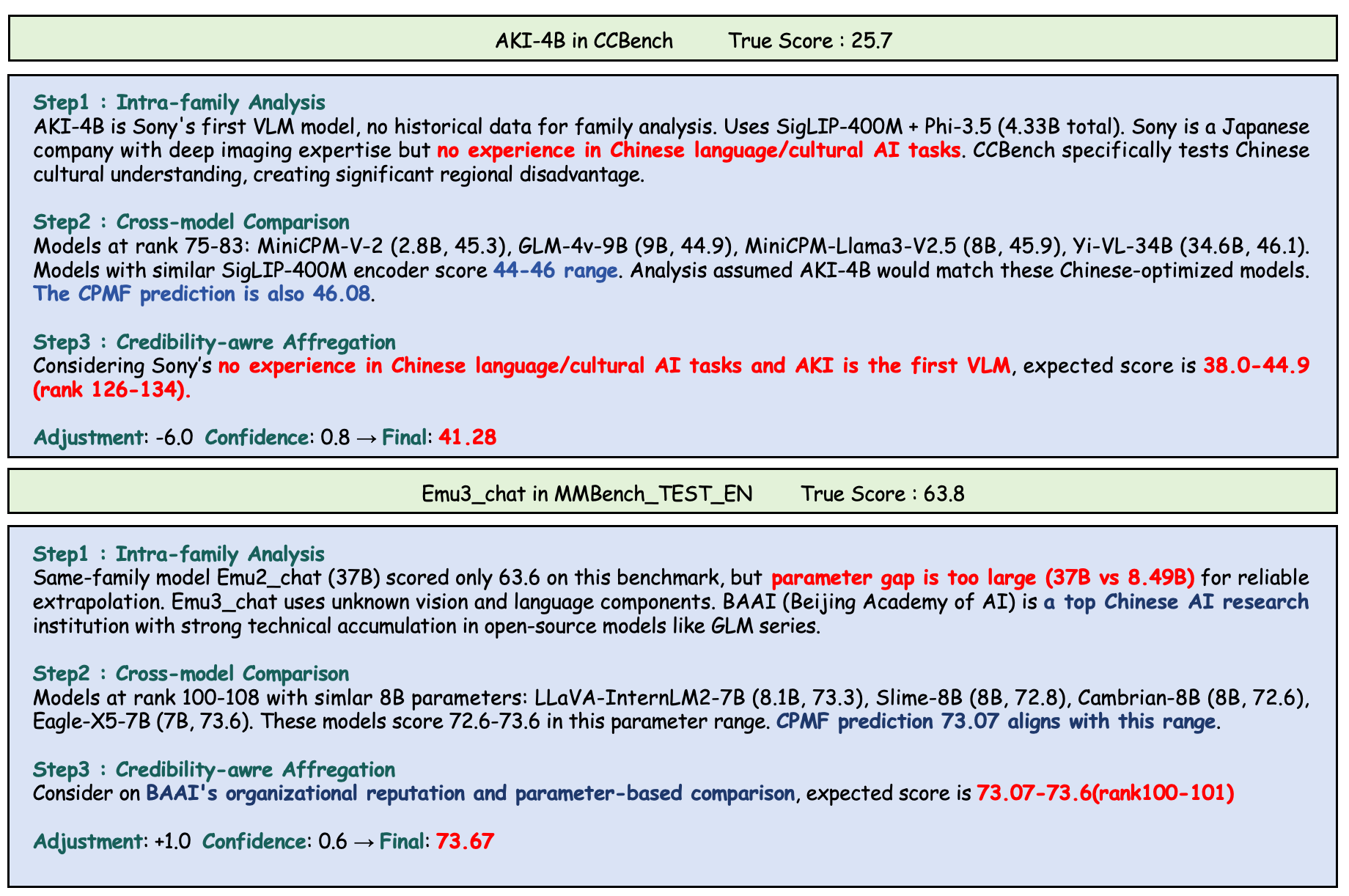}
    \caption{Failed prediction cases. STAR produces large errors of 15.58 and 9.87 for AKI-4B on CCBench and Emu3\_chat on MMBench\_TEST\_EN respectively. The reasoning process identifies relevant factors but fails to accurately quantify their impact.}
    \label{fig:bad_case}
\end{figure}

\textbf{Analysis of Failed Cases.}
The first failure reveals limitations in quantifying cultural-linguistic gaps. While STAR correctly identifies that Sony has no experience in Chinese language/cultural AI tasks and applies a negative adjustment of -6.0, this correction proves insufficient as the true score of 25.7 is far below the predicted 41.28. The cross-model comparison misleadingly references Chinese-optimized models like MiniCPM-V-2 and GLM-4v-9B, which have fundamentally different training data distributions. The second failure illustrates the danger of parameter-based comparisons. STAR assumes Emu3\_chat should perform similarly to other 8B models scoring 72-73, but the actual score of 63.8 suggests Emu3\_chat may use different architecture or training approaches that parameter count alone cannot capture. Notably, Emu2\_chat at 37B scored only 63.6, which could have served as a better reference despite the parameter gap. These cases highlight that STAR's reasoning can identify relevant factors but may struggle to accurately estimate their quantitative impact, particularly for models with unique characteristics or extreme distribution shifts.

\section{Real-world Evaluation}
\label{app:real_word}
We evaluate three selection strategies (Brute-force Evaluation, Random Selection, and STAR-guided Selection) over a pool of \(285\) candidate models and a single Math benchmark.

 For each model \(m \in \mathcal{C}\) and benchmark \(b \in \mathcal{B}_{\text{math}}\), let \(s_{m,b}\) be the true OpenCompass score of \(m\) on \(b\).
We define the ground-truth Math score of model \(m\) as
\[
f(m) = \frac{1}{|\mathcal{B}_{\text{math}}|}
\sum_{b \in \mathcal{B}_{\text{math}}} s_{m,b},
\]
which reduces to the single Math score in our setting.
Using \(f(m)\), we obtain the ground-truth top‑10 set
\[
T = \operatorname{Top10}_{m \in \mathcal{C}} f(m),
\]
i.e., the ten models with the highest ground-truth Math scores.

Under a given evaluation budget, each strategy \(S\) produces a recommended set of ten models \(R_S\).
We report the following metric:

\textbf{Top‑10 Recall.}
For each strategy \(S\), we define the Top‑10 Recall as
\[
\operatorname{Top10Recall}(S) = \frac{\lvert R_S \cap T \rvert}{10},
\]
which measures the fraction of ground-truth top‑10 models that are successfully recovered by strategy \(S\).

For Random Selection, we further consider three evaluation budgets corresponding to evaluating \(25\%\), \(50\%\), and \(75\%\) of the model–benchmark pairs.
For each budget level \(p \in \{0.25, 0.50, 0.75\}\), we run the random strategy five times with different seeds and report the average Top‑10 Recall over these five runs.

\begin{table}[h]
  \centering
  \caption{Comparison of selection strategies in terms of normalized evaluation cost and Top‑5 Recall on 285 models and one Math benchmark. For Random Selection, we report the average over five runs at three evaluation budgets.}
  \label{tab:top5_recall_cost}
  \begin{tabular}{lc}
    \toprule
    Strategy &  Top‑10 Recall \\
    \midrule
    Brute-force Evaluation  (100\% budget)             & 1.00  \\
    Random Selection (25\% budget)      & 0.24  \\
    Random Selection (50\% budget)      & 0.52  \\
    Random Selection (75\% budget)      & 0.70  \\
    STAR-guided Selection     (3.5\% budget)          & 0.82  \\
    \bottomrule
  \end{tabular}
\end{table}

As shown in Table.\ref{tab:top5_recall_cost}, with only $3.5\%$ of the full evaluation budget, our method achieves a Top-10 Recall of $0.82$, 
outperforming random selection even under a $75\%$ evaluation budget. 
This demonstrates the strong practicality of our approach in real-world settings, 
allowing researchers to drastically reduce evaluation costs while still quickly identifying the best models.

\section{Supplementary Results for Ablation Experiment}
\label{app:ablation}

\subsection{Ablation on STAR Componment}
\label{app:ablation_com}
Table~\ref{tab:app_ablation} presents the ablation study that isolates the contribution of each component in STAR. We organize the ablation into two parts: statistical components and semantic adjustment modules.

\begin{table*}[h]
\centering
\small
\setlength{\tabcolsep}{4.5pt}
\begin{tabular}{lcccccccc}
\toprule
 \multirow{2}{*}{\textbf{Method}} & \multicolumn{3}{c}{\textbf{Score-based Metrics$\downarrow$}} 
 & \multicolumn{4}{c}{\textbf{Rank-based Metrics$\uparrow$}} 
 & \multicolumn{1}{c}{\textbf{Total$\uparrow$}} \\
\cmidrule(lr){2-4} \cmidrule(lr){5-8} \cmidrule(lr){9-9}
& RMSE & MAE & Score Avg 
& SRCC(\%) & KRCC(\%) & MAE@3(\%) & Rank Avg(\%) 
& Score \\
\midrule

\rowcolor{gray!15}\multicolumn{9}{l}{\textit{Ablation on Statistical Components}} \\

PMF (w/o features) 
& 10.82 & 7.18 & 9.00 
& 94.46 & 86.42 & 34.48 & 71.79 
& 62.79 \\

\quad + Model Features 
& 10.21 & 6.78 & 8.50 
& 94.89 & 86.98 & 34.12 & 72.00 
& 63.50 \\

\quad + Benchmark Features (CPMF) 
& 9.95 & 6.47 & 8.21 
& 95.27 & 87.56 & 33.94 & 72.26 
& 64.05 \\

\midrule
\rowcolor{gray!15}\multicolumn{9}{l}{\textit{Ablation on Semantic Adjustment}} \\

CPMF + LLM Direct 
& 9.68 & 6.31 & 8.00 
& 95.01 & 87.24 & 34.52 & 72.26 
& 64.26 \\

\quad + Family Evolution 
& 9.42 & 6.12 & 7.77 
& 95.38 & 87.68 & 35.41 & 72.82 
& 65.05 \\

\quad + Rank-Similar Models 
& 9.15 & 5.92 & 7.54 
& 95.72 & 88.12 & 36.85 & 73.56 
& 66.02 \\

\quad + Reasoning (STAR) 
& \textbf{8.75} & \textbf{5.69} & \textbf{7.22} 
& \textbf{96.10} & \textbf{88.64} & \textbf{38.23} & \textbf{74.32} 
& \textbf{67.10} \\

\bottomrule
\end{tabular}

\caption{Ablation study on STAR components. We progressively add components: (1) \textbf{Statistical}: model and benchmark features improve PMF baseline; (2) \textbf{Semantic Adjustment}: LLM Direct provides basic adjustment, Family Evolution adds same-series performance analysis, Rank-Similar Models adds cross-family comparison, and Reasoning enables explicit chain-of-thought for final prediction.}

\label{tab:app_ablation}
\end{table*}

\textbf{Statistical Components.} Starting from the vanilla PMF baseline, we progressively incorporate side information. Adding model features (parameters, organization) reduces the score-based average from 9.00 to 8.50, demonstrating that model metadata provides useful inductive bias. Further incorporating benchmark features (category, difficulty) yields CPMF with an average of 8.21, confirming that benchmark characteristics help capture task-specific performance patterns.

\textbf{Semantic Adjustment.} Building upon CPMF, we examine the impact of LLM-based refinement. The ``LLM Direct'' variant, where the LLM adjusts predictions without structured evidence, provides marginal improvement (8.21$\to$8.00), indicating that naive LLM intervention has limited value. Introducing \textit{Family Evolution} evidence---which analyzes performance trends within the same model series---further reduces the average to 7.77. Adding \textit{Rank-Similar Models} evidence, which compares the target model against similarly-ranked models from training data, yields additional gains (7.54). Finally, enabling explicit chain-of-thought \textit{Reasoning} produces the full STAR system with the best performance (7.22), highlighting the importance of structured reasoning over direct score prediction.

\subsection{Ablation on LLM Backbone}
\label{app:ablation_backbone}

Table~\ref{tab:app_llm_backbone} examines the impact of different LLM backbones on STAR's performance. Several findings emerge from this analysis. First, all LLM-augmented variants outperform the CPMF baseline, with Total Score improvements ranging from 0.48 to 3.05 points. This validates that EVT-guided reasoning consistently enhances prediction quality regardless of the specific LLM used. Second, closed-source frontier models achieve comparable performance, with GPT-5.1 at 67.10, Claude-3.5-Sonnet at 66.90, and Gemini-2.5-Pro at 66.72. The marginal differences of less than 0.4 points suggest that STAR's structured reasoning framework effectively leverages the capabilities of various frontier models. Third, open-source models provide competitive alternatives. Qwen-2.5-72B achieves a Total Score of 65.56, only 1.54 points below GPT-5.1, demonstrating that STAR can be deployed with fully open-source components while maintaining strong performance. Even the lightweight Qwen-2.5-7B scores 64.53, outperforming CPMF by 0.48 points despite having significantly fewer parameters. These results indicate that STAR's performance gains primarily stem from its reasoning framework design rather than raw model capability, making it adaptable to various deployment constraints including cost-sensitive or privacy-requiring scenarios.

\begin{table*}[h]
\centering
\small
\setlength{\tabcolsep}{4.5pt}
\begin{tabular}{lcccccccc}
\toprule
\multirow{2}{*}{\textbf{LLM Backbone}} & \multicolumn{3}{c}{Score-based Metrics$\downarrow$} & \multicolumn{4}{c}{Rank-based Metrics$\uparrow$} & \multicolumn{1}{c}{Total$\uparrow$} \\
\cmidrule(lr){2-4} \cmidrule(lr){5-8} \cmidrule(lr){9-9}
& RMSE & MAE & Score Avg & SRCC(\%) & KRCC(\%) & MAE@3(\%) & Rank Avg(\%) & Score \\
\midrule

CPMF (no LLM) 
& 9.95 & 6.47 & 8.21 
& 95.27 & 87.56 & 33.94 & 72.26 
& 64.05 \\

\midrule
\rowcolor{gray!15}\multicolumn{9}{l}{\textit{Closed-Source Models}} \\

GPT-5.1 
& \textbf{8.75} & \textbf{5.69} & \textbf{7.22} 
& \textbf{96.10} & \textbf{88.64} & \textbf{38.23} & \textbf{74.32} 
& \textbf{67.10} \\

Claude-3.5-Sonnet 
& 8.82 & 5.74 & 7.28 
& 96.02 & 88.51 & 38.02 & 74.18 
& 66.90 \\

Gemini-2.5-Pro 
& 8.88 & 5.78 & 7.33 
& 95.95 & 88.42 & 37.78 & 74.05 
& 66.72 \\

\midrule
\rowcolor{gray!15}\multicolumn{9}{l}{\textit{Open-Source Models}} \\

Qwen-2.5-72B 
& 9.28 & 6.08 & 7.68 
& 95.62 & 88.05 & 36.04 & 73.24 
& 65.56 \\

Qwen-2.5-7B 
& 9.72 & 6.38 & 8.05 
& 95.35 & 87.72 & 34.68 & 72.58 
& 64.53 \\

\bottomrule
\end{tabular}

\caption{Impact of LLM backbone on STAR performance. Closed-source frontier models achieve comparable results, while open-source Qwen-2.5-72B provides competitive performance. Even the lightweight Qwen-2.5-7B outperforms the CPMF baseline.}
\label{tab:app_llm_backbone}
\end{table*}

\subsection{Ablation on Retrieval Sources}
\label{app:ablation_retrival}

\begin{table*}[h]
\centering
\small
\setlength{\tabcolsep}{4.5pt}
\begin{tabular}{lcccccccc}
\toprule
\multirow{2}{*}{\textbf{Retrieval Sources}} 
& \multicolumn{3}{c}{Score-based Metrics$\downarrow$} 
& \multicolumn{4}{c}{Rank-based Metrics$\uparrow$} 
& \multicolumn{1}{c}{Total$\uparrow$} \\
\cmidrule(lr){2-4} \cmidrule(lr){5-8} \cmidrule(lr){9-9}
& RMSE & MAE & Score Avg 
& SRCC(\%) & KRCC(\%) & MAE@3(\%) & Rank Avg(\%) 
& Score \\
\midrule

CPMF (no LLM) 
& 9.95 & 6.47 & 8.21 
& 95.27 & 87.56 & 33.94 & 72.26 
& 64.05 \\

\midrule

All Sources (Full) 
& \textbf{8.75} & \textbf{5.69} & \textbf{7.22} 
& \textbf{96.10} & \textbf{88.64} & \textbf{38.23} & \textbf{74.32} 
& \textbf{67.10} \\

\midrule

w/o HuggingFace 
& 9.15 & 5.98 & 7.57 
& 95.72 & 88.15 & 36.52 & 73.46 
& 65.89 \\

w/o arXiv 
& 9.02 & 5.86 & 7.44 
& 95.88 & 88.35 & 37.18 & 73.80 
& 66.36 \\

w/o Google 
& 8.92 & 5.78 & 7.35 
& 95.98 & 88.48 & 37.65 & 74.04 
& 66.69 \\

\midrule

Only HuggingFace 
& 9.42 & 6.18 & 7.80 
& 95.52 & 87.88 & 35.28 & 72.89 
& 65.09 \\

Only arXiv 
& 9.72 & 6.42 & 8.07 
& 95.25 & 87.58 & 34.15 & 72.33 
& 64.26 \\

Only Google 
& 9.88 & 6.55 & 8.22 
& 95.08 & 87.38 & 33.62 & 72.03 
& 63.81 \\

\midrule

No Retrieval 
& 10.35 & 6.92 & 8.64 
& 94.68 & 86.95 & 31.85 & 71.16 
& 62.52 \\

\bottomrule
\end{tabular}

\caption{Contribution of different retrieval sources under 95\% masking. Without retrieval, the LLM-based adjustment performs substantially worse than the full STAR setting, demonstrating that structured evidence is essential for effective semantic refinement.}
\label{tab:app_retrieval_sources}
\end{table*}
Table~\ref{tab:app_retrieval_sources} analyzes the contribution of different retrieval sources to STAR's performance. The results reveal several important findings. First, combining all retrieval sources achieves the best performance at 67.10, outperforming any single-source configuration by at least 1.21 points. This demonstrates the complementary nature of different information sources. Second, among individual sources, HuggingFace contributes most significantly. Removing HuggingFace causes the largest performance drop from 67.10 to 65.89, while removing arXiv or Google results in smaller decreases to 66.36 and 66.69 respectively. This aligns with expectations, as HuggingFace model cards provide structured technical specifications directly relevant to performance prediction. Third, when using only a single source, HuggingFace alone achieves 65.09, outperforming arXiv-only at 64.26 and Google-only at 63.81. Notably, all single-source configurations still outperform the No Retrieval setting at 62.52. Fourth, the No Retrieval configuration performs worse than even CPMF at 64.05, indicating that LLM reasoning without structured evidence can introduce harmful noise rather than beneficial adjustments. This validates our design choice of retrieval-augmented reasoning: the LLM's role is to interpret and synthesize retrieved evidence rather than generate predictions from parametric knowledge alone.

\section{Runtime and Computational Overhead}
\label{app:runtime}

We report the runtime and computational overhead of STAR on the OpenCompass setup to assess its practicality as a drop-in tool for leaderboard maintenance and model selection.

\subsection{Offline Amortized Cost}

The statistical backbone of STAR is the semantic-augmented CPMF model described in Section~\ref{sec:statistical}. We train CPMF once per dataset and reuse it for all subsequent predictions. In our experiments with \(M = 284\) models and \(N = 28\) benchmarks, we run NUTS with 100 warmup and 100 sampling iterations, yielding \(T = 100\) posterior samples as used in Section~\ref{sec:statistical}. On a single CPU server, this training completes in approximately 8--12 minutes.

The runtime of CPMF scales roughly linearly with the number of observed entries \(\lvert \mathbf{R}^{\mathrm{obs}} \rvert\) and the latent dimension \(D\). For the relatively small matrix in our experiments, CPMF training is two to three orders of magnitude cheaper than running a single full evaluation pass of all multimodal benchmarks for one large model.

Semantic features for models and benchmarks are encoded once using BGE-M3 and stored in a feature database. Loading these features and the CPMF parameters into memory at the beginning of an evaluation session takes less than one second and is negligible compared to LLM inference.

\subsection{Online Prediction Cost}

Table~\ref{tab:runtime} summarizes the online cost of producing one prediction for a given model–benchmark pair. The critical-path operations consist of a fast database lookup followed by three short LLM calls in the EVT-guided reasoning module.

\begin{table}[h]
\centering
\caption{Wall-clock runtime of STAR components on the OpenCompass evaluation setup. Offline CPMF training is amortized across all predictions; online cost is dominated by three LLM calls.}
\label{tab:runtime}
\begin{tabular}{lcc}
\toprule
\textbf{Component} & \textbf{Time} & \textbf{Frequency} \\
\midrule
\multicolumn{3}{l}{\textit{Offline / amortized}} \\
CPMF training (NUTS, 100+100 iters) & \(\approx 8\text{--}12\) min & once per dataset \\
Feature database loading & \(< 1\) s & once per session \\
\midrule
\multicolumn{3}{l}{\textit{Online / per prediction}} \\
Evidence lookup (DB + index) & \(< 0.1\) s & per model–benchmark pair \\
Family-evolution analysis (LLM) & 2--4 s & per model–benchmark pair \\
Rank-similar comparison (LLM) & 2--4 s & per model–benchmark pair \\
EVT-based adjustment (LLM) & 2--4 s & per model–benchmark pair \\
\midrule
\textbf{Total online latency} & \textbf{6--12 s} & per prediction \\
\bottomrule
\end{tabular}
\end{table}

\paragraph{LLM inference.}
STAR makes three LLM calls per prediction: (1) \emph{family evolution}, which analyzes performance trends within the same model family; (2) \emph{rank-similar comparison}, which contrasts the target model with CPMF-predicted peer models; and (3) \emph{EVT-based adjustment}, which aggregates the evidence and produces the final calibrated prediction and explanation. In our setup, each call processes approximately 800--1{,}200 input tokens and generates 150--250 output tokens. Using GPT-5.1 via the OpenAI API, we observe a latency of 2--4 seconds per call, resulting in a total online latency of 6--12 seconds per prediction as reported in Table~\ref{tab:runtime}.

\paragraph{Batch evaluation.}
Predictions for different model–benchmark pairs are independent and can be parallelized without communication. In the 95\% masking experiment, we require on the order of \(10^3\) predictions; running them sequentially takes roughly 3--4 hours. With \(K\) parallel workers, the wall-clock time decreases approximately linearly, e.g., to below 15 minutes with \(K=16\).

\subsection{Cost and Scalability}

\paragraph{Monetary cost.}
Assuming pricing comparable to GPT-4o (2.50 USD per million input tokens and 10.00 USD per million output tokens), the token usage above (around 3{,}000 input and 500--750 output tokens per prediction) corresponds to roughly 0.01--0.02 USD per prediction. For a typical leaderboard update involving 10 new models on 28 benchmarks (280 predictions), the total API cost is therefore on the order of 3--5 USD.

\paragraph{Scalability optimizations.}
STAR supports several straightforward strategies to further reduce latency and cost:

\begin{itemize}[noitemsep,leftmargin=*]
\item \textbf{Parallel calls.} The three LLM calls per prediction target disjoint subtasks and can be issued in parallel, which reduces the end-to-end latency to the maximum of the three (about 2--4 seconds).
\item \textbf{Batch parallelization.} Multiple model–benchmark pairs can be evaluated concurrently using standard data-parallel execution, achieving near-linear speedup with the number of workers.
\item \textbf{Cheaper backbones.} As shown in Table~\ref{tab:llm_backbone}, replacing GPT-5.1 with smaller open-source models such as Qwen-2.5-7B yields Total scores that are still higher than the CPMF-only baseline, while reducing per-call cost by an order of magnitude.
\item \textbf{Local deployment.} Deploying a strong open-source model (e.g., Qwen-2.5-72B) locally removes per-call API charges and allows STAR to run at cluster throughput limited only by local hardware.
\item \textbf{Caching.} Evidence retrieval and intermediate analysis for frequently queried model–benchmark pairs can be cached, avoiding repeated LLM reasoning when the underlying CPMF posterior has not changed.
\end{itemize}

\paragraph{Comparison with full evaluation.}
Compared to existing evaluation workflows, STAR introduces modest additional computation while substantially reducing GPU-bound benchmarking cost. Running the full OpenCompass suite for a single large multimodal model typically requires hours of GPU time, whereas STAR amortizes a one-time CPU-only CPMF training cost of around 10 minutes and then answers each new model–benchmark query in a few seconds using LLM calls. This trade-off makes STAR particularly attractive for scenarios such as rapid leaderboard updates, pre-screening candidates before full evaluation, and estimating performance in settings where running all benchmarks is infeasible.

Overall, STAR strikes a practical balance between prediction quality and computational overhead. Offline CPMF training is lightweight and amortized over many predictions, while online latency of 6--12 seconds per query is acceptable for non-real-time model assessment. When combined with parallelization and cheaper LLM backbones, STAR scales to thousands of predictions at a cost that is negligible compared to running full benchmark suites or relying on human experts.

\section{List of Prompts}
This section details the specific prompts utilized to guarantee experimental reproducibility. To support open research, we will also make the retrieval logs of STAR publicly available. Moreover, we aim to deploy STAR as a gratuitous service in the future, facilitating rapid model selection for the LLM development community.

\begin{figure*}[t]
    \centering
    \includegraphics[width=\textwidth]{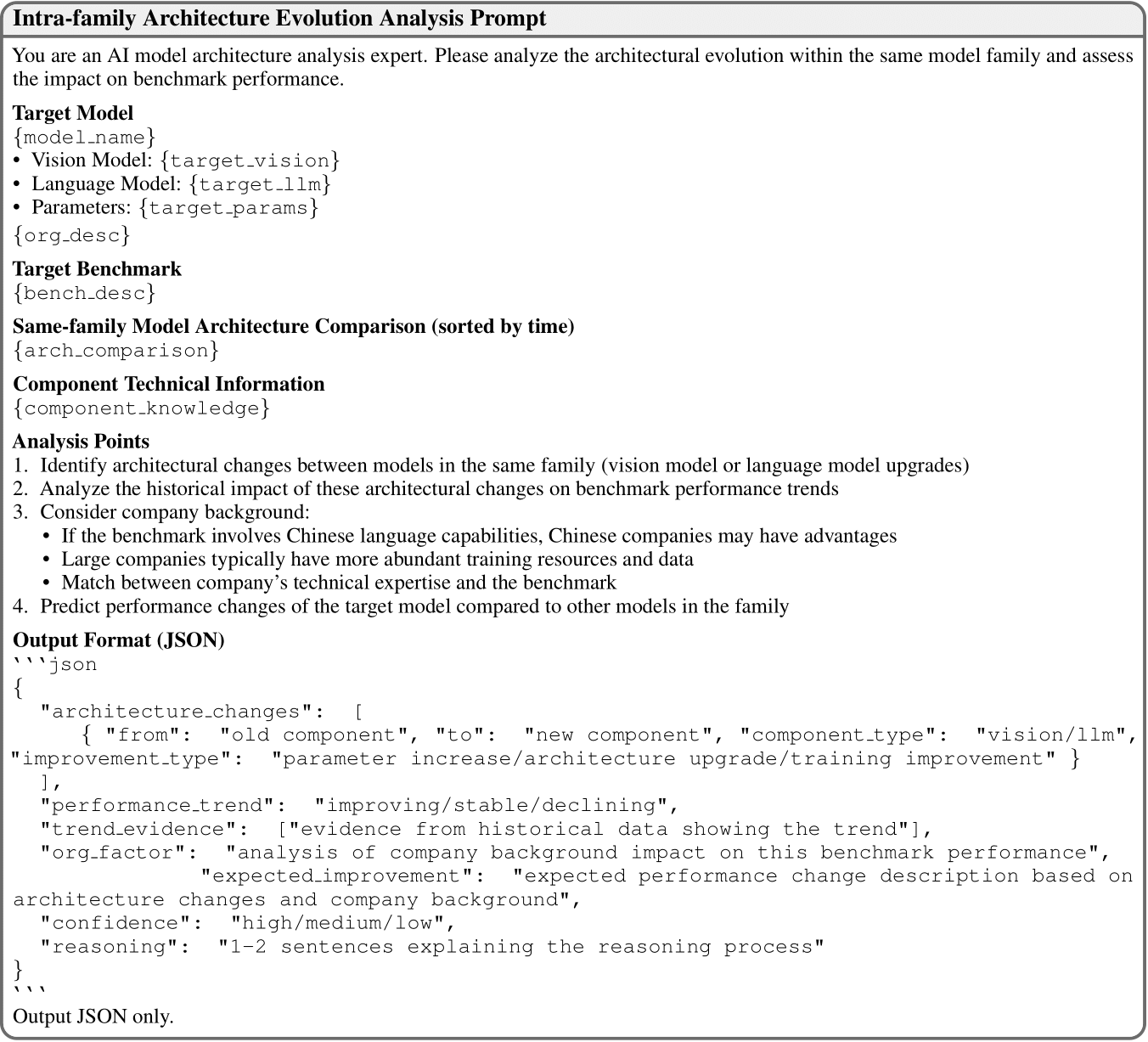}
    
\end{figure*}

\begin{figure*}[t]
    \centering
    \includegraphics[width=\textwidth]{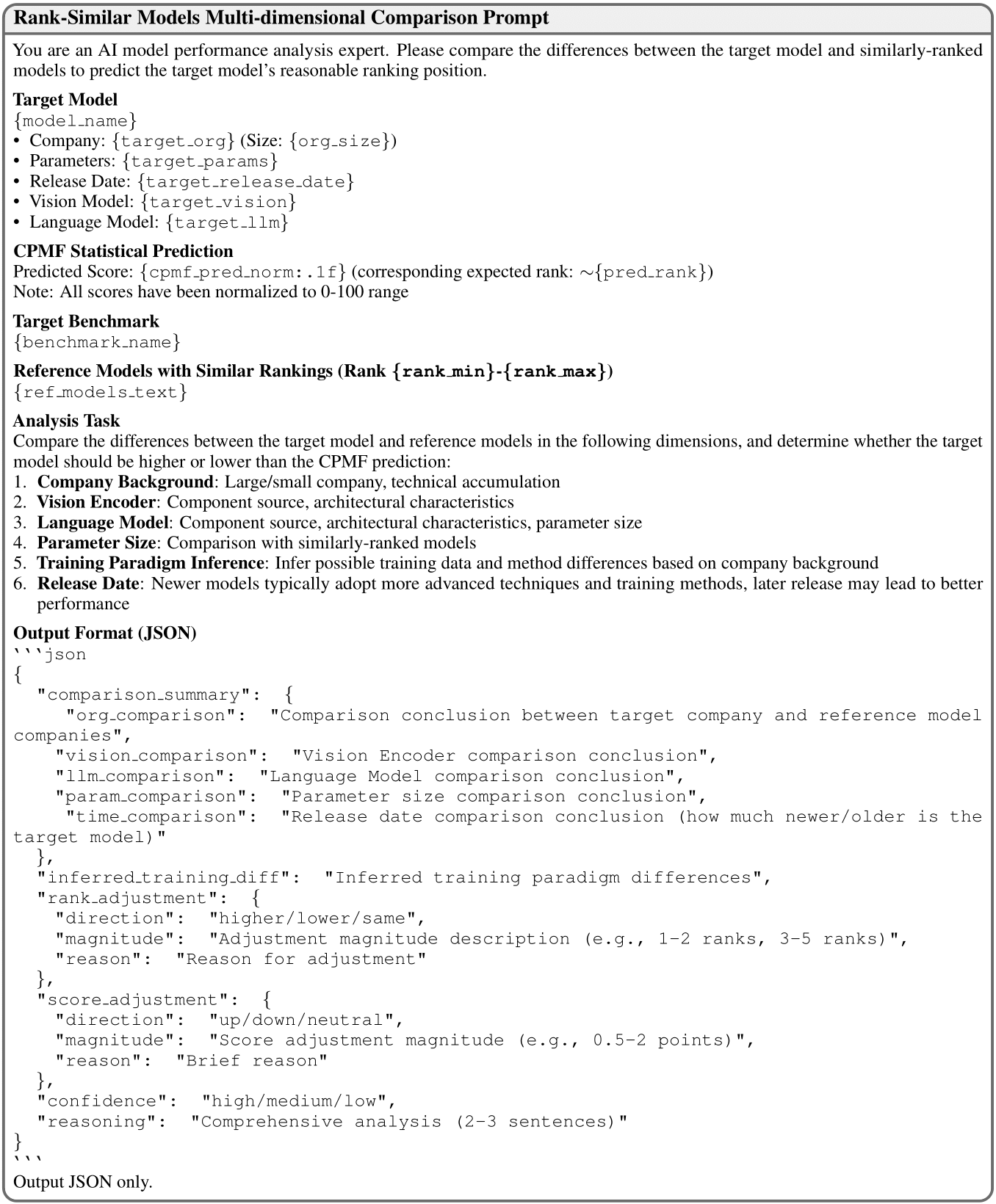}
    
\end{figure*}

\begin{figure*}[t]
    \centering
    \includegraphics[width=\textwidth]{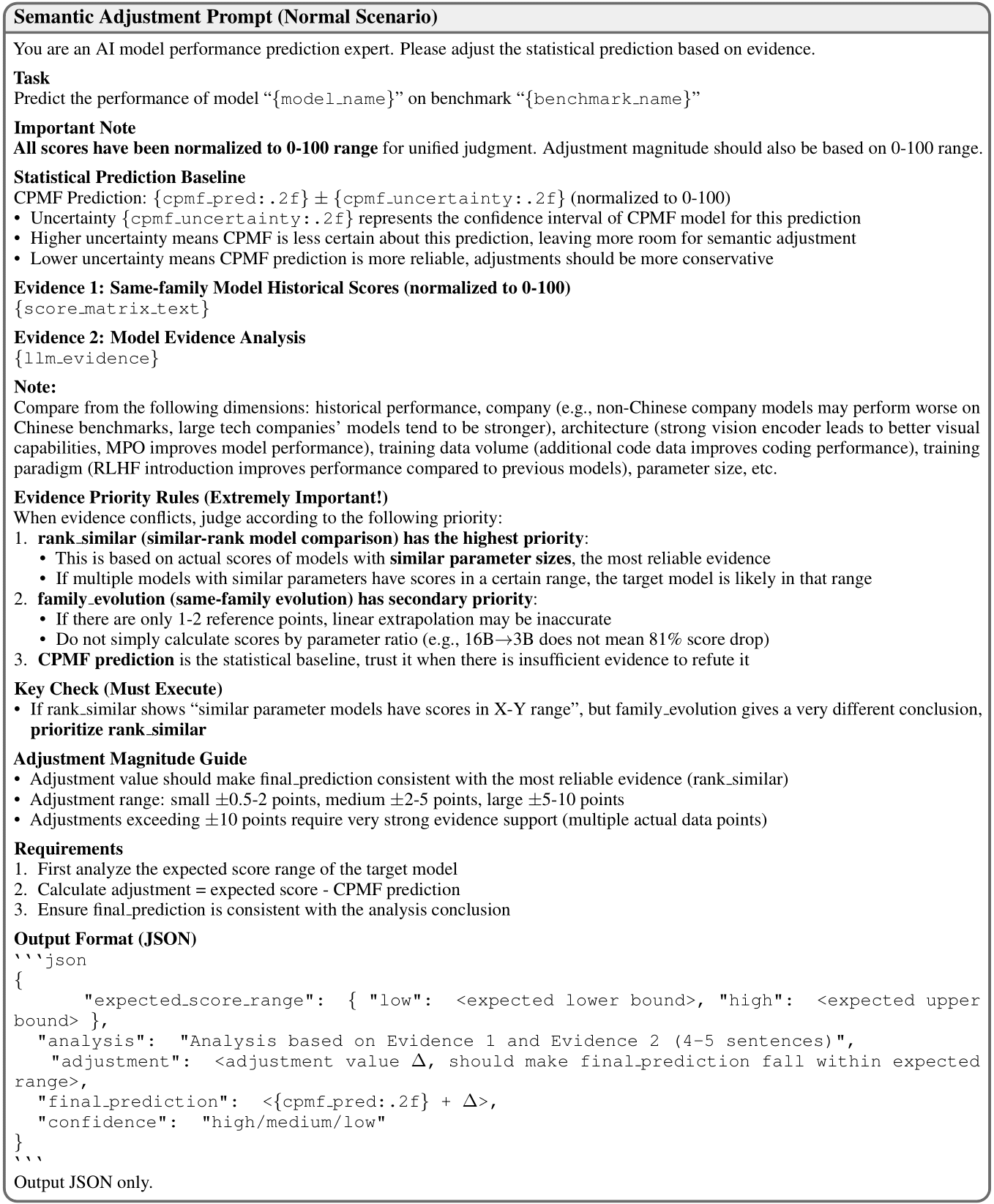}
    
\end{figure*}

\begin{figure*}[t]
    \centering
    \includegraphics[width=\textwidth]{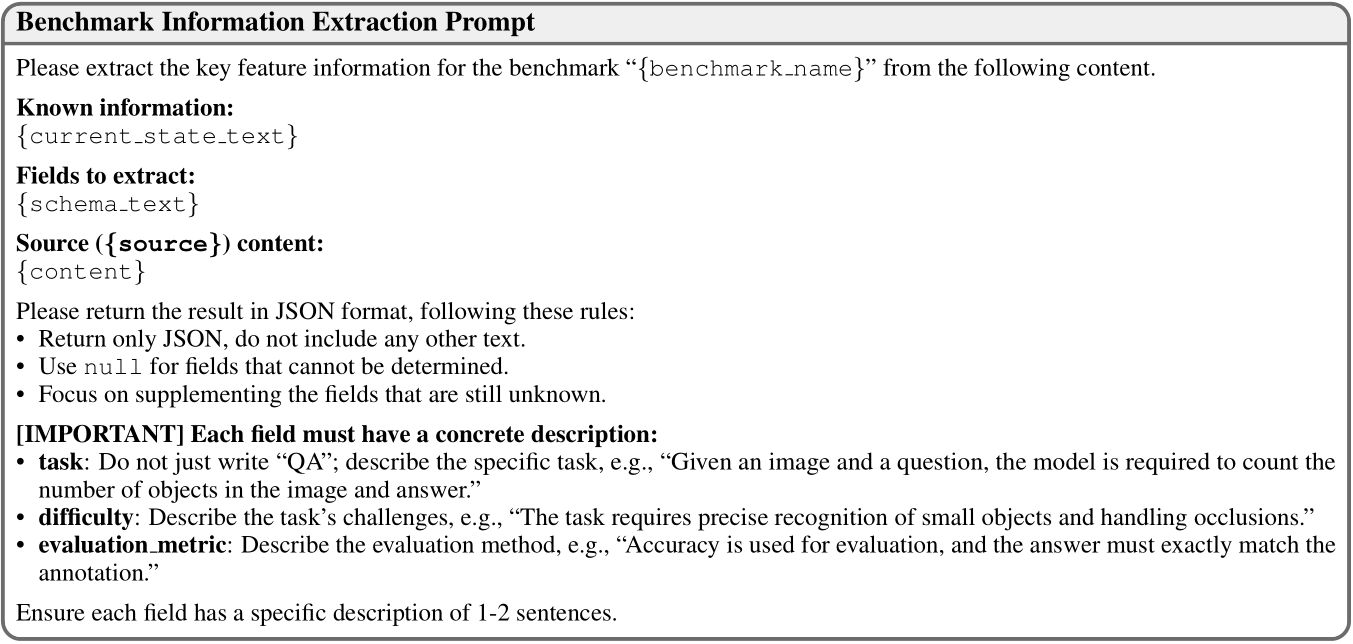}
    
\end{figure*}

\begin{figure*}[t]
    \centering
    \includegraphics[width=\textwidth]{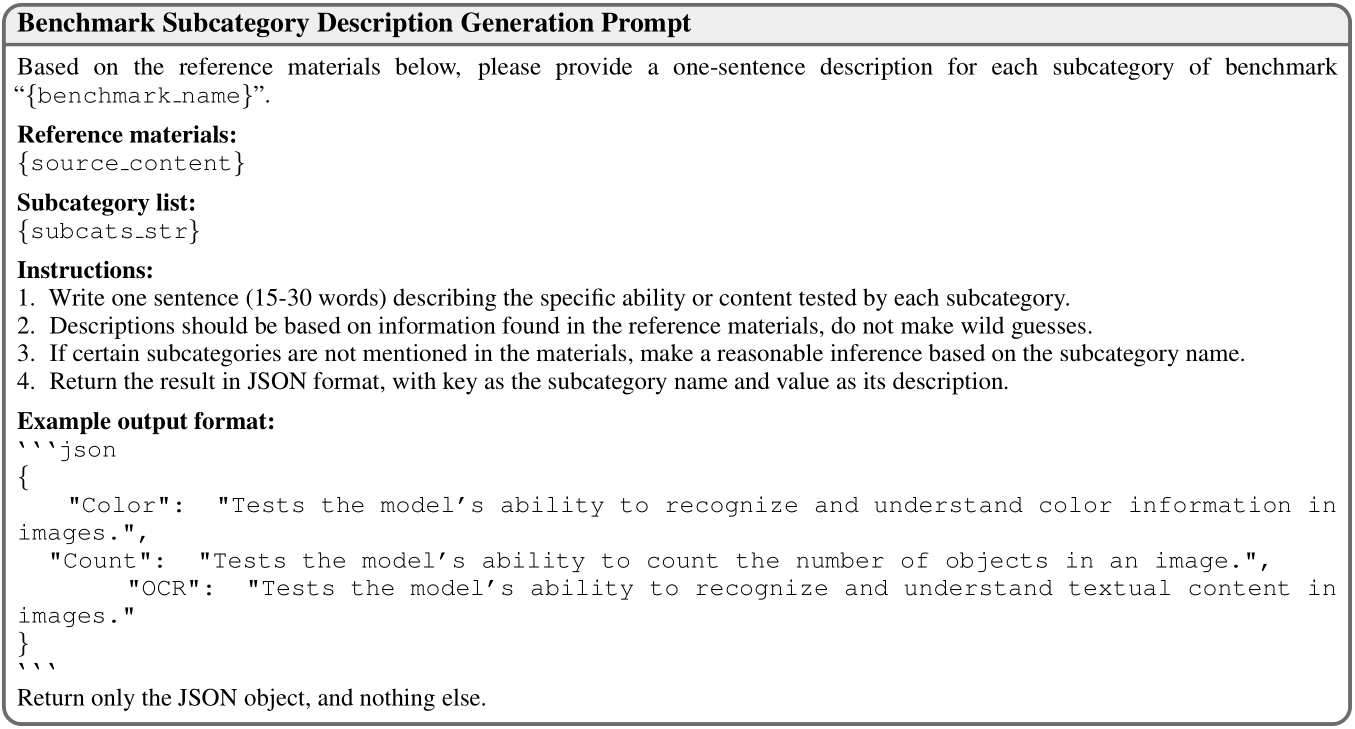}
    
\end{figure*}

\begin{figure*}[t]
    \centering
    \includegraphics[width=\textwidth]{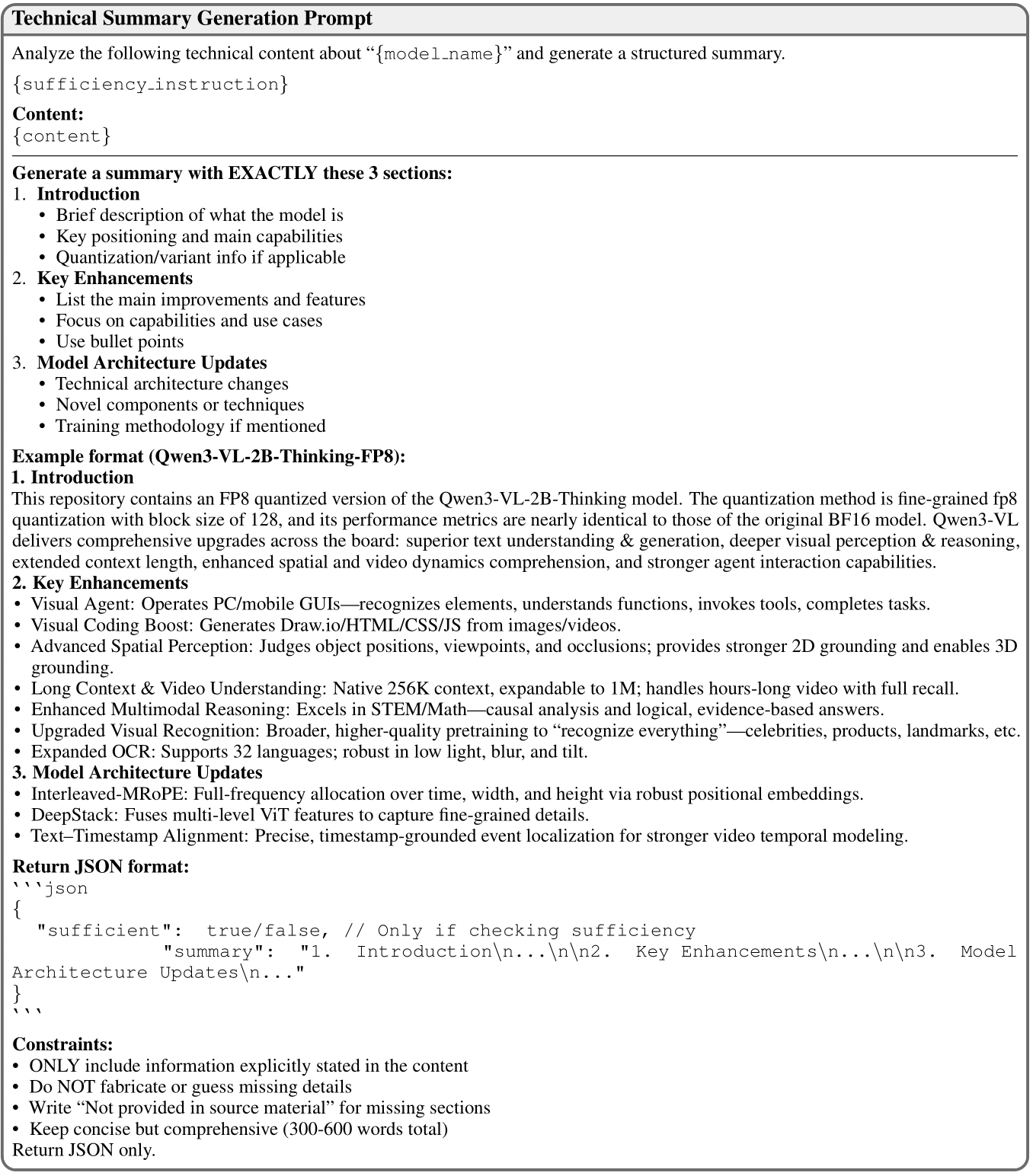}
    
\end{figure*}

\begin{figure*}[t]
    \centering
    \includegraphics[width=\textwidth]{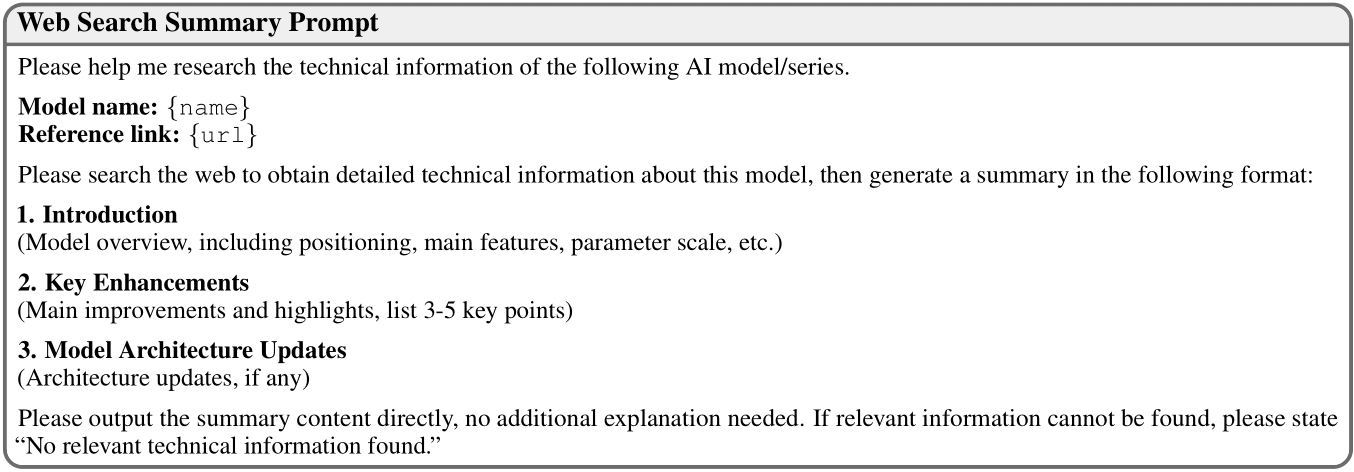}
    
\end{figure*}

\begin{figure*}[t]
    \centering
    \includegraphics[width=\textwidth]{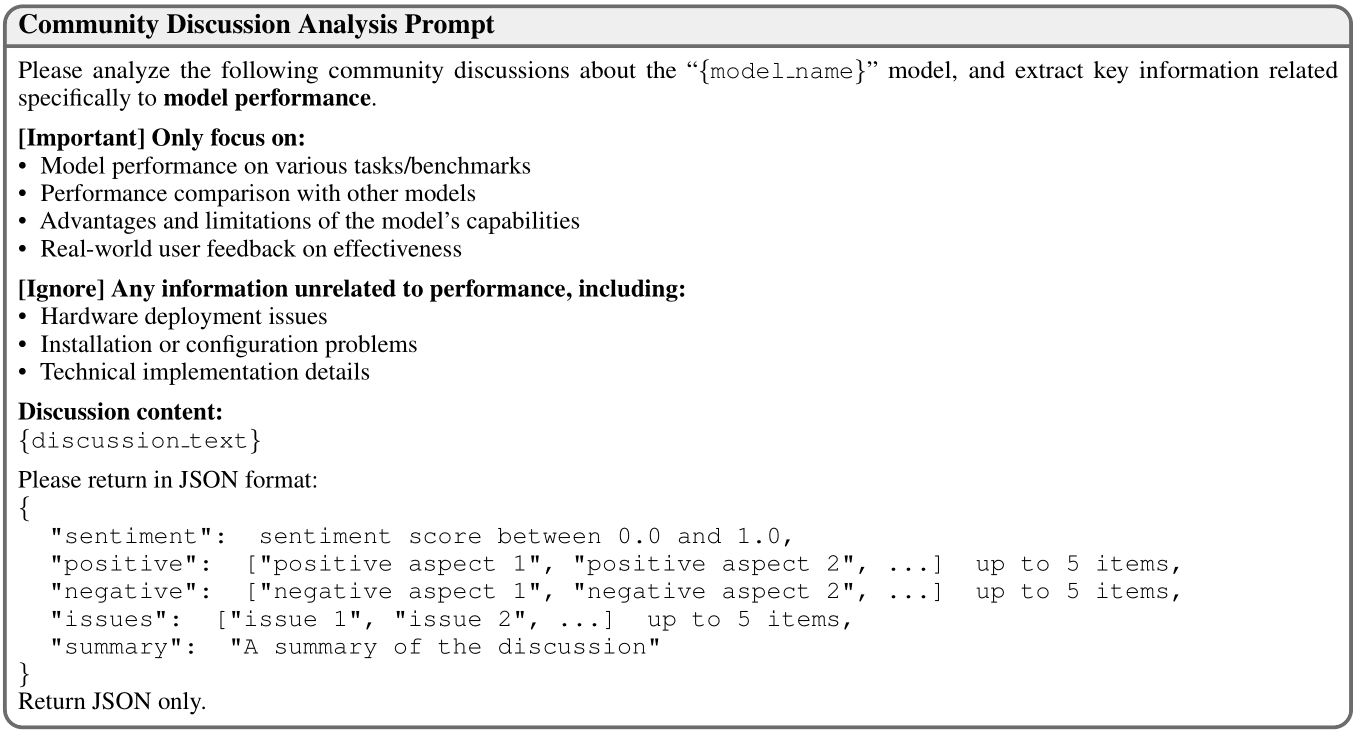}
    
\end{figure*}


\end{document}